\documentclass[10pt,twocolumn]{article}
\usepackage[margin=0.75in,columnsep=0.25in]{geometry}
\usepackage[T1]{fontenc}
\usepackage{times}
\usepackage{graphicx}
\usepackage{booktabs}
\usepackage{multirow}
\usepackage{amsmath,amssymb}
\usepackage{xcolor}
\usepackage[hidelinks]{hyperref}
\usepackage{caption}
\usepackage{tikz}
\usetikzlibrary{arrows.meta,positioning,calc,fit}
\usepackage{pgfplots}
\pgfplotsset{compat=1.17}

\title{\textbf{Continuously Evolving Deepfake Detection:\\
An Architecture and Public-Benchmark Evaluation of a\\
Dynamic Detection System}}
\author{Ken Jon Miyachi\\
BitMind\\
\texttt{ken@bitmind.ai}
\and
Dylan Uys\\
BitMind\\
\texttt{dylan@bitmind.ai}}
\date{}

\begin{document}

\makeatletter
\twocolumn[
  \@maketitle
  \begin{center}
  \begin{minipage}{0.86\textwidth}
  \begin{center}\textbf{Abstract}\end{center}
  \small
Deepfake detectors that achieve near-perfect scores on academic benchmarks
collapse on real-world content: recent in-the-wild evaluations report AUC
drops of 45--50\% for state-of-the-art open-source models. We argue this
gap is structural: static detectors are trained once against a moving
generative frontier. We present BitMind Forensics (BMF), trained through
Bittensor SN34, an open adversarial competition that continually refreshes
the training distribution. We evaluate one dated export comprising image,
general-video, and human-video checkpoints across nineteen public datasets:
the canonical face-swap suites (FaceForensics++, Celeb-DF v1/v2/++, DFDC,
DFD, UADFV, DF40) and recent in-the-wild and AI-generated-media benchmarks (Sumsub,
Deepfake-Eval-2024, WildRF, Community Forensics, AIGCDetectBench,
GenImage, AI-GenBench, AIGIBench, RAID, GenVidBench, GenVideo-100K).
BMF reaches 0.936 AUC on the Sumsub benchmark's original images and
0.872 pooled AUC over its full four-condition manipulation battery
(1,407,414 images), remaining effective under manipulation (0.855 under
JPEG, 0.799 under downscaling), while GPEN enhancement improves detection
(0.996). On Deepfake-Eval-2024, it matches the best commercial
detector on images (0.915 vs 0.90) and exceeds it on video (0.822 vs
0.79), far above the best open-source detectors (0.56 and 0.63). It reaches
0.991 AUC on a 21-generator AI-image panel and 0.918 on GenVidBench,
and exceeds the FF++-trained frontier on DFDC (0.947 vs 0.843) and
Celeb-DF v2 (0.9985 vs 0.956), both contamination-audited, with
statistical parity on Celeb-DF++. In a temporal study, successive dated
exports improve on held-out media from generators absent from the
static baseline's training (image 0.842 to 0.902; video 0.864 to 0.936).
Our evaluation harness is public, and at publication the production API
serves the exact evaluated snapshot for independent verification.
  \end{minipage}
  \end{center}
  \vspace{1.4em}
]
\makeatother

\section{Introduction}

The defining problem of deepfake detection is not accuracy but
staleness: detectors that dominate the benchmarks of their training era
collapse on the content of the next one. State-of-the-art detectors report cross-dataset AUCs above
0.95 on canonical face-forgery evaluations~\cite{effort}, and
near-perfect accuracy on million-scale synthetic-image
suites~\cite{genimage}. Yet when the same detectors meet content that
actually circulates, such as deepfakes collected from social media,
fraudulent identity-verification attempts, and political disinformation,
performance collapses. On Deepfake-Eval-2024, a benchmark of real
deepfakes circulated during 2024, open-source state-of-the-art models
lose 45--50\% of their AUC relative to academic benchmarks, and no
commercial detector tested reached 90\% accuracy~\cite{deepfakeeval2024}.
On a perceptually hard test set of community-generated images, the best
published method achieves roughly 65\% accuracy~\cite{aide}. Fewer than
half of the open-source detectors evaluated against modern face-swap
pipelines exceed 0.60 AUC, and simple manipulations (JPEG compression,
downscaling, off-the-shelf face enhancement) are sufficient to push
several below chance~\cite{sumsub}. Audits of deployed commercial tools
reach the same conclusion on real political deepfakes~\cite{fitforpurpose,
publictools}. The gap between benchmark and real-world performance is not
a margin of \mbox{error}; it is the dominant fact
about the field.
We argue this gap is structural rather than incidental. A deepfake
detector is trained once, against a fixed corpus of forgeries, and then
deployed into an environment where the generative frontier moves weekly:
new diffusion architectures, new face-swap pipelines, new commercial
video generators, and post-processing pipelines that scrub away forensic
traces all enter circulation faster than any annual benchmark cycle can
capture. Detection is therefore a
\emph{non-stationary} adversarial problem, and a static model, however
well it generalizes at training time, begins decaying the moment its
training distribution is frozen. Recent longitudinal studies confirm
this directly: detectors trained on pre-2023 data degrade measurably on
fakes from 2024--2025~\cite{decay}, and benchmarks have begun
incorporating temporal protocols that train on early generators and test
on later ones precisely because chronological drift, not architectural
weakness, drives most observed failure~\cite{communityforensics}. The
implication is uncomfortable for the standard paradigm: the question is
not only \emph{which model} detects fakes, but \emph{which process keeps
a detector current}.

This paper presents BitMind Forensics (BMF), a deepfake detection
system built around that second question. Rather than training a single
model against a fixed dataset, BMF is built on the outputs of an open
incentive mechanism in which independent participants
(\emph{miners}) compete on both sides of the problem: generative miners
serve state-of-the-art generators, producing a continuously refreshed
stream of verified synthetic media, and discriminative miners submit
detector checkpoints that compete to classify it. The mechanism
functions as a form of online adversarial training at the system level:
as new generators appear, they enter the challenge distribution within
hours (validators refresh the challenge set on a roughly four-hour
cycle), and the detector population is economically compelled to adapt
or lose reward. The deployed detector, a dated export comprising a
1.11B-parameter image ensemble and separate video checkpoints
(\S\ref{sec:arch}), is derived from this process as a discrete
snapshot: miners surface detector architectures and adversarial
training data, and we fine-tune, extend, and temporalize the winning
basis into the production checkpoints.
The central design claim is that \emph{the system adapts even though every
individual snapshot is static}: any frozen export inherits a training
distribution that is days, not years, behind the generative frontier.

We evaluate one dated BMF export (April~15, 2026), without per-benchmark tuning,
across public benchmarks spanning face-swap deepfakes, AI-generated
images, and AI-generated video, under both clean and manipulated
(compressed, downscaled, enhancement-laundered) conditions. On the
Sumsub in-the-wild robustness benchmark~\cite{sumsub}, our snapshot
achieves 0.936 AUC on original images, improving on the strongest prior
baseline for every generator, and 0.872 pooled AUC over the full
$1{,}407{,}414$-image manipulation battery
(\S\ref{sec:sumsub}). On the
Deepfake-Eval-2024 in-the-wild image track it reaches 0.915 AUC, and
on a 21-generator AI-image panel it reaches 0.991 AUC; it also
generalizes zero-shot across face-swap datasets it was never trained on
(\S\ref{sec:face}). On AI-generated video, its general video model
reaches 0.918 AUC on GenVidBench and 0.822 AUC on the
Deepfake-Eval-2024 video track, and on the canonical face-swap
cross-dataset benchmarks its human-video specialist \emph{exceeds} the
FF++-trained specialist Effort on DFDC and Celeb-DF~v2, with parity on
Celeb-DF++, on contamination-audited
subsets (\S\ref{sec:face-video}).
In a temporal study, successive dated exports improve on a fixed test set
of media from generators absent from the static baseline's training, on
both the image track ($0.842 \to 0.902$, a gain of
$6.0$ AUC points over a November snapshot) and the video track
($0.864 \to 0.936$, $+7.2$ points) (\S\ref{sec:temporal}).

Because we evaluate our own product, we report TPR at fixed
low false-positive rates alongside threshold-free metrics, evaluate the
exact snapshot our production API serves, and use a publicly released
evaluation harness. These operating-point numbers are not flattering
everywhere (at $1\%$ FPR, video-track recall on Deepfake-Eval-2024 is
$24\%$; \S\ref{sec:analysis}), and we report them directly rather than
leading with AUC alone.

Our contributions are:
\begin{enumerate}
  \item \textbf{Architecture and mechanism.} We describe an open,
  incentive-driven training process for deepfake detection and the
  heterogeneous ensemble built on it (\S\ref{sec:arch}), framing
  continuous data refresh as the primary defense against generative
  drift.
  \item \textbf{Benchmark evaluation.} We evaluate one fixed
  snapshot across nineteen public datasets under a single protocol with
  no per-benchmark adaptation: the in-the-wild suites Sumsub,
  Deepfake-Eval-2024, and WildRF; the AI-generated-image suites Community
  Forensics, AIGCDetectBench, GenImage, AI-GenBench, and AIGIBench; the
  adversarial-robustness suite RAID; the face-deepfake cross-dataset
  canon (FaceForensics++, UADFV, DFD, DFDC, DF40, Celeb-DF v1/v2/++); and
  the AI-generated-video suites GenVidBench and GenVideo-100K
  (\S\ref{sec:protocol}--\ref{sec:results}). We compare against the
  strongest published result available under each benchmark's reported
  protocol, making protocol mismatches explicit where they remain
  (\S\ref{sec:vsbest}).
  \item \textbf{Robustness analysis.} We quantify performance under the
  manipulation attacks that defeat prior detectors (JPEG, downscaling,
  enhancement laundering), and analyze calibration and low-FPR operating
  points relevant to production moderation (\S\ref{sec:sumsub},
  \S\ref{sec:analysis}).
  \item \textbf{GAS-Station.} We release GAS-Station, an adversarial dataset
  of AI-generated images ($50{,}399$, from $78$ generative participants)
  and video produced by the Subnet~34 competition: media generated
  specifically to fool the current detector frontier
  (\S\ref{sec:protocol}).
  \item \textbf{Adaptation over time.} We show that successive dated
  snapshots track the generative frontier: on media from generators
  absent from a static baseline's training, AUC improves with snapshot
  recency on both the image and video
  tracks (\S\ref{sec:temporal}).
\end{enumerate}

\section{Related Work}\label{sec:related}

\subsection{Deepfake and AIGC Detection}
Early deepfake detection treated the problem as supervised binary
classification over known forgery types, exemplified by Xception
baselines trained on FaceForensics++~\cite{ffpp}. Because such models
overfit to method-specific artifacts, a second generation pursued
generalization through artifact synthesis and representation design:
self-blended images simulate boundary artifacts without using any real
forgeries~\cite{sbi}, CADDM counters identity leakage with multi-scale
local artifact modeling~\cite{caddm}, and latent-space augmentation
widens the forgery decision boundary~\cite{lsda}. The current frontier
is dominated by adapting large pretrained vision backbones, on the
evidence that representation quality drives cross-domain transfer:
linear probes on frozen CLIP features outperform dedicated
architectures~\cite{unifd}, frequency-aware and prompt-based CLIP
adapters add forgery sensitivity~\cite{fatformer,c2pclip}, and
parameter-efficient subspace methods such as Effort achieve the
strongest published cross-dataset results on face benchmarks while
preserving the backbone's general knowledge~\cite{effort}. For
fully synthetic imagery, complementary signal families include
diffusion reconstruction error~\cite{dire,drct}, upsampling traces in
neighboring pixel relations~\cite{npr}, mixture-of-experts over
low-level and semantic features~\cite{aide}, and bias-controlled
training data construction~\cite{bfree}. Most recently, multimodal
LLMs have been applied both as zero-shot detectors, where they trail
specialized models but offer explanations~\cite{gpt4mf}, and as
fine-tuned detection assistants~\cite{fakeshield}. Our work is
orthogonal to this axis of architectural progress: we hold the
\emph{training process} responsible for generalization, and treat the
detector architecture as a component that the process continually
re-equips with current training data.

\subsection{Benchmarks and the Generalization Gap}
Evaluation infrastructure has consolidated around standardized
benchmarks. DeepfakeBench unified preprocessing and protocols across
classic face-forgery datasets~\cite{deepfakebench}, and DF40 extended
coverage to 40 modern generation methods spanning face-swap,
reenactment, and diffusion-based synthesis~\cite{df40}. For generated
images, GenImage scaled evaluation to millions of samples across eight
generators~\cite{genimage}, though later analysis showed its results
are confounded by compression and resolution shortcuts~\cite{fakeorjpeg},
a finding that motivated bias-controlled successors. A parallel line
of work measures performance where it matters most: on content that
circulates in the wild. Deepfake-Eval-2024 collected deepfakes shared
on social media during 2024 and found that open-source detectors lose
roughly half their AUC, with no commercial system reaching 90\%
accuracy~\cite{deepfakeeval2024}; the Chameleon test set of
human-deceptive community images drives state-of-the-art detectors
toward chance~\cite{aide}; and the Sumsub benchmark shows that simple
quality manipulations defeat most open-source
detectors~\cite{sumsub}. Audits of deployed commercial tools on real
political deepfakes~\cite{fitforpurpose} and blinded evaluations of
public detection platforms~\cite{publictools} corroborate the gap.
Generated video is the newest frontier, with million-scale corpora
covering commercial text-to-video systems~\cite{genvideo,genvidbench}.
We draw our evaluation suite from this literature, prioritizing
in-the-wild and manipulation-robust protocols over saturated academic
sets, and we evaluate against the published numbers of each
benchmark's strongest baselines.

\subsection{Keeping Detectors Current}
The community has begun treating temporal drift as a first-class
evaluation axis. Longitudinal studies show detectors trained on
pre-2023 corpora decay measurably on 2024--2025
forgeries~\cite{decay}, AI-GenBench formalizes a sliding-window
protocol that trains on generators released before a cutoff and tests
on those released after~\cite{aigenbench}, and Community Forensics
demonstrates that generalization scales with the number and diversity
of training generators, assembling data from thousands of community
models~\cite{communityforensics}. These results imply that data
freshness and generator coverage, not architecture alone, determine
real-world performance. Existing responses are largely manual:
periodic dataset releases, challenge cycles, and ad hoc retraining.
Our system differs in making refresh \emph{mechanistic}: an open system
incentivizes new data and adversarial examples, so the training process
tracks the generative frontier without depending on any single curator,
and the cost of staying current is distributed across competing
participants.
The mechanism we evaluate builds on our earlier work introducing the
BitMind generative-adversarial architecture, a framework
that uses economic incentives to drive continual innovation in deepfake
detection~\cite{sotf}; the present paper freezes a single production
snapshot of that system and subjects it to extensive independent
benchmarking. To our knowledge, no other published detection system
couples training to an open incentive mechanism of this kind.

\section{System Architecture}\label{sec:arch}
BMF comprises three detectors built on a shared foundation of pretrained
vision backbones (Table~\ref{tab:branches}): an
\emph{image model} (a 1.11B-parameter heterogeneous ensemble operating
directly on raw RGB pixels), a \emph{general video model} (the same
backbones temporalized with per-branch temporal adapters), and a
\emph{human-video model} (a face-cropped specialist for face-centric
deepfake video that uses the same underlying backbones and adds
video-native and frequency-domain branches).
Requests dispatch to one of the three by modality and content; this
section describes the image model first (\S\ref{sec:ensemble}--\ref{sec:input}),
then both video models (\S\ref{sec:video-arch}), and finally the
incentive loop that surfaces their architecture basis and continually
refreshes the training distribution of all three
(\S\ref{sec:incentive}).

The image model was not designed from scratch: its core adopts the
architecture of round-winning miner submissions accumulated across
successive SN34 competition rounds (each round is winner-take-all,
\S\ref{sec:incentive}), a three-branch ensemble
(ConvNeXt-L~\cite{convnext}, EVA-L~\cite{eva02}, and CLIP
ViT-L~\cite{clip}; branches A1, A2, and B in
Table~\ref{tab:branches}) whose task heads we further fine-tuned and
which we extended with a DINOv3 ViT-L branch (C)~\cite{dinov3}. The competition thus serves as an architecture-search
and data engine at once: miners surface what works, and the production
model builds on it. All
image results in this paper use the base ensemble path (384$^2$ input,
softmax $P(\text{fake})$) so that the evaluated configuration matches
the production (``organics'') inference contract used for the rescored
benchmarks, i.e., benchmarks whose metrics we recompute from raw media
through our own inference pipeline rather than citing published
numbers.
The central design commitment is heterogeneity: by combining
backbones that disagree in architecture family, pretraining objective,
and input resolution, we make it less likely that any single
generator-specific shortcut survives ensemble averaging.

\subsection{Neural Ensemble}\label{sec:ensemble}
The detector comprises four classifiers spanning three pretraining
paradigms and three native resolutions (Table~\ref{tab:branches}). Each
branch bilinearly resizes the shared input to its own native scale,
applies its own normalization, and emits two logits
$[\ell_{\text{real}}, \ell_{\text{fake}}]$.

\begin{table}[t]
\centering\small
\setlength{\tabcolsep}{4pt}
\caption{The four ensemble branches. Effective fusion weights are
$\tfrac{1}{6}, \tfrac{1}{6}, \tfrac{1}{3}, \tfrac{1}{3}$ for
A1, A2, B, C respectively (see fusion equations).}
\label{tab:branches}
\begin{tabular}{llll}
\toprule
Br. & Backbone & Pretraining & Input \\
\midrule
A1 & ConvNeXt-L (CLIP soup) & CLIP $\to$ IN-12k/1k & 384$^2$ \\
A2 & EVA-L/14              & MIM $\to$ IN-22k/1k  & 336$^2$ \\
B  & CLIP ViT-L/14         & Vision--language     & 224$^2$ \\
C  & DINOv3 ViT-L/16       & Self-supervised      & 224$^2$ \\
\bottomrule
\end{tabular}
\end{table}

\paragraph{Heads.}
Branches A1 and A2 use bounded cosine classifiers: features and weights
are $L_2$-normalized, scaled by $s=30$, passed through
$15\tanh(\cdot/15)$, and given a fixed bias of $[0, +0.5]$ on the fake
logit. Branch B applies dropout ($0.3$) to the CLS token followed by a
linear map to two logits. Branch C concatenates the pooled CLS token
with the mean of the patch tokens (2048-d) and applies an MLP
($2048\!\to\!512\!\to\!2$), with output order permuted to match the
ensemble convention. The $\tanh$-bounded cosine heads of A1 and A2 cap
per-branch confidence, which (together with logit averaging) discourages
saturation toward hard labels; the calibration this actually yields is
measured, not assumed (\S\ref{sec:calib}, \S\ref{sec:analysis}).

\paragraph{Fusion.}
Fusion is uniform logit averaging at two levels. The two
ImageNet-style branches are first pooled, then combined with the
contrastive and self-supervised branches:
\begin{align}
g_A(x) &= \tfrac{1}{2}\bigl(g_{\text{cnx}}(x) + g_{\text{eva}}(x)\bigr),\\
f_{\text{neural}}(x) &= \tfrac{1}{3}\bigl(g_A(x) + g_{\text{clip}}(x)
   + g_{\text{dino}}(x)\bigr),
\end{align}
giving effective per-branch weights of $\tfrac{1}{6}$ (A1), $\tfrac{1}{6}$
(A2), $\tfrac{1}{3}$ (B), and $\tfrac{1}{3}$ (C). We deliberately mix
ConvNet and ViT families, contrastive, MIM, and self-supervised
objectives, and resolutions from 224 to 384 px, on the hypothesis that
generator-specific artifacts learned by one representation are weakly
correlated with those learned by the others and therefore attenuate
under averaging. The final prediction is
$P(\text{fake}) = \operatorname{softmax}(f_{\text{neural}}(x))_{\text{fake}}$,
the softmax mass on the fake logit.

\subsection{Input Pipeline}\label{sec:input}
Each of the three models has a fixed input contract, held constant
across every benchmark it scores. The \emph{image model} takes a single
\texttt{uint8} RGB frame, center-oriented and resized or cropped to
$384\times384$ by a deterministic upstream stage (the gasbench/organics
preprocessor), with no augmentations applied at inference; each
branch is otherwise self-contained over the raw pixels, performing its
own internal resize and normalization (ImageNet statistics for A1, A2,
and C; CLIP statistics for B). The \emph{general video model} takes
$8$-frame clips at the same $384^2$ full-frame resolution, with
identical per-branch handling applied per frame. The \emph{human-video
model} is the one path with content-dependent preprocessing: an MTCNN
face detector crops the face region with a $1.3\times$ margin (falling
back to a center crop when no face is found) before the temporal stack.

For the image and general-video paths, the absence of any
face-detection or cropping stage is deliberate: they score the full
frame. This is a meaningful departure from several academic protocols,
including the Sumsub baselines, which preprocess with RetinaFace or
dlib face crops before classification. For those benchmarks we report
full-frame scores and note the protocol difference as a limitation
(\S\ref{sec:limitations}) rather than adopting a benchmark-specific
crop stage. The human-video specialist's face crop is likewise a
uniform, domain-level choice, fixed across all benchmarks it scores.
The evaluation claim is therefore one configuration \emph{per modality
path}, never tuned per benchmark.

Deployment routing is by \emph{modality} at the API layer, not by
per-image expert gating. Every image request executes all four branches;
video and audio requests dispatch to separate checkpoints
(dedicated video models, \S\ref{sec:video-arch}, and a WavLM audio
model, out of scope here). Optional domain overrides, such as
customer-specific weights, lazy-load from a domain directory and fall
back to the default production model when absent.

No post-hoc calibration is applied anywhere in the system\label{sec:calib}:
scores are raw softmax $P(\text{fake})$ at temperature $1.0$, with no
Platt or temperature scaling. The selection objective
(\S\ref{sec:incentive}) does include a Brier-score term, so probabilistic
accuracy is rewarded during training, but we make no calibration claim
beyond what \S\ref{sec:analysis} measures: expected calibration error is
reasonable on clean and near-distribution data and degrades substantially
under heavy manipulation and recompression.

\subsection{Video Pipeline}\label{sec:video-arch}
Video requests are served by two separate checkpoints, not the image
ensemble, and both build on the same underlying backbone families
(Table~\ref{tab:branches}). Like the image model, both are in-house
builds on the competition's outputs: we temporalize and extend the
image ensemble ourselves on the competition data stream. The \emph{general video model}
($\approx$1.12B parameters) temporalizes the image ensemble's backbones:
each branch runs per-frame on $8$-frame $384^2$ inputs and receives a
multi-scale temporal adapter (parallel depthwise 1-D convolutions over
the time axis at kernel sizes $3/5/7$ with a residual bottleneck), branch
logits are combined by a \emph{learned} softmax-weighted fusion (in
contrast to the image model's fixed averaging), and an attention module
pools per-frame features into the video-level logit. The
\emph{human-video model} ($\approx$1.45B parameters), used for
face-centric deepfake video, is a temporalized fork of the image
ensemble: its branches and heads are initialized from the production
image weights, temporal pooling is upgraded to $8$-head cross-attention
over frames, and two additional branches are fused at a second level: a
video-native self-supervised V-JEPA~2 encoder~\cite{vjepa2} (the only component in the
system that models motion natively rather than per-frame) and a
frequency-domain DCT branch. Unlike the full-frame image and
general-video paths, the human-video model operates on face-cropped
inputs (MTCNN detection~\cite{mtcnn} with $1.3\times$ margin and a center-crop
fallback), a domain-level preprocessing choice applied uniformly, not
per benchmark. At inference,
\texttt{ffmpeg} decodes the input and samples a fixed budget of
approximately $16$ frames, which are grouped into temporal chunks; each
chunk yields a clip-level $P(\text{fake})$, and the chunk scores are
aggregated by confidence weighting so that high-certainty segments
dominate the video-level decision. For longer inputs, a ``full'' mode
tiles the clip with overlapping temporal windows and pools their scores,
trading compute for coverage of intermittent or temporally localized
manipulations. All video results in this paper use the standard
fixed-budget mode; the full tiling mode is not used in any reported
evaluation. Because the video and image checkpoints are distinct, the
image results that form the bulk of this paper do not depend on the video
path; video benchmarks are reported separately in \S\ref{sec:video}.

\subsection{Incentive Mechanism}\label{sec:incentive}
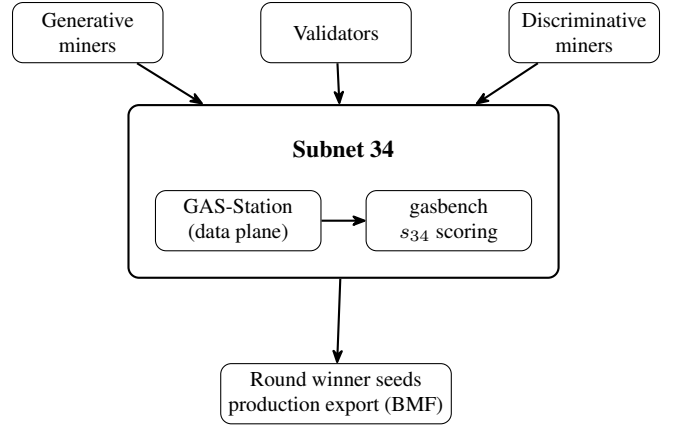
\begin{figure}[t]
  \centering
  \resizebox{\linewidth}{!}{%
  \begin{tikzpicture}[
    font=\footnotesize,
    >={Stealth[round]},
    actor/.style={draw, rounded corners, align=center, minimum
      height=8mm, minimum width=20mm, inner sep=3pt, fill=white},
    data/.style={draw, rounded corners, align=center, minimum
      height=8mm, minimum width=22mm, inner sep=3pt, fill=white},
    prod/.style={draw, rounded corners, align=center, minimum
      height=8mm, minimum width=24mm, inner sep=3pt, fill=white},
    arr/.style={->, thick},
  ]
    \node[actor] (gen) at (-0.4,0) {Generative\\miners};
    \node[actor] (val) at (2.9,0) {Validators};
    \node[actor] (disc) at (6.2,0) {Discriminative\\miners};
    \node[data] (gas) at (1.6,-2.5) {GAS-Station\\(data plane)};
    \node[data] (bench) at (4.4,-2.5) {gasbench\\$s_{34}$ scoring};
    \node[font=\small\bfseries] (snlabel) at (3.0,-1.55) {Subnet 34};
    \node[draw, rounded corners, thick, fit=(gas)(bench)(snlabel),
      inner sep=10pt] (subnet) {};
    \node[prod] (org) at (2.9,-4.8) {Round winner seeds\\production export (BMF)};

    \draw[arr] (gen) -- (subnet);
    \draw[arr] (val) -- (subnet);
    \draw[arr] (disc) -- (subnet);
    \draw[arr] (gas) -- (bench);
    \draw[arr] (subnet) -- (org);
  \end{tikzpicture}}
  \caption{The SN34 loop. Generative miners, validators, and
  discriminative miners interact through Subnet~34: verified
  miner-generated media are versioned in the GAS-Station data plane,
  and detector submissions are scored by the $s_{34}$ objective in
  gasbench runs over GAS-Station data, public benchmarks, and a
  private holdout. Each round's winning basis seeds the production
  export (BMF), which we fine-tune and extend before it serves the
  organics detection API (\S\ref{sec:ensemble}).}
  \label{fig:system}
\end{figure}

BMF builds on an ongoing
competition run as Bittensor Subnet~34 (SN34): its architecture basis
and its training distribution are products of that competition, which
we fine-tune and extend into the production checkpoints
(\S\ref{sec:ensemble}, \S\ref{sec:video-arch}). Three classes of actors
participate (Figure~\ref{fig:system}). \emph{Generative miners}
contribute synthetic media by serving state-of-the-art generation
models; \emph{validators} issue prompts to the generative miners and
verify each response with strict C2PA-metadata and prompt-alignment
checks; and \emph{discriminative miners} submit detector checkpoints
that compete to classify media. Verified miner-generated media are
versioned in GAS-Station, and detector scoring runs as periodic
evaluations of our harness, \texttt{gasbench}, over a combination of
public benchmarks, a private holdout, and fresh GAS-Station data.
Generative miners additionally earn a \emph{delayed} reward tied to the
rate at which their media fools the current detector population, so the
challenge distribution is continually pushed toward what the deployed
frontier fails to detect.

Each detector submission receives a scalar reward that credits both
discrimination and calibration. Writing $\widetilde{\mathrm{MCC}}$ for the
normalized Matthews correlation coefficient and $\widetilde{B}$ for a
normalized, inverted Brier score (so that larger is better for both), the
per-round score is their weighted geometric mean,
\begin{equation}
s_{34} \;=\; \Bigl(\widetilde{\mathrm{MCC}}^{\,\alpha}\cdot
\widetilde{B}^{\,\beta}\Bigr)^{\!1/(\alpha+\beta)},
\qquad \alpha = 1.2,\ \beta = 1.8 .
\end{equation}
The geometric form means a detector cannot win by maximizing raw
discrimination while emitting miscalibrated probabilities: a weak Brier
term multiplies the whole score down. Reward is \emph{winner-take-all} per
modality per round, so the single best discriminator captures that round's
emissions and is adopted as the basis of the production export that
serves the (``organics'') detection API (\S\ref{sec:arch}).

The mechanism's defining property is that its challenge distribution is
\emph{endogenous and adversarial}. Because generative miners earn the
delayed fool-rate reward for media that the current detector population
fails to flag, and are therefore economically driven to serve the newest
released generators and configurations, the training and evaluation
distribution tracks the generative frontier without a human curator. This is, in effect, online adversarial
training at the level of an open market: every snapshot exported from the
system, although individually frozen, inherits a data distribution that is
days rather than years behind the latest generators. This is the central
claim this paper sets out to test. We treat the current winner-take-all
tournament as the live mechanism; richer ``king-of-the-hill'' schemes that
retain a stable of complementary specialists across rounds are left to
future work.

\section{Evaluation Protocol}\label{sec:protocol}
\subsection{Model Under Test}
All main results come from a single dated BMF export (April~15, 2026),
comprising the image, general-video, and human-video checkpoints described
in \S\ref{sec:arch}, with no per-benchmark tuning. The temporal study
(\S\ref{sec:temporal}) also evaluates earlier exports, and the ablation
(\S\ref{sec:ablations}) decomposes the image checkpoint into its branches.
Model weights are not public (see the Reproducibility Statement). At
publication, the production API serves this snapshot using the same model
and code as our local evaluation. The decision threshold is
fixed at $0.5$ on each checkpoint's softmax $P(\text{fake})$ output and held
constant across all benchmarks; no per-dataset or post-hoc threshold
calibration is performed.

\subsection{Benchmarks}
We draw the evaluation suite from the in-the-wild and manipulation-robust
benchmarks surveyed in \S\ref{sec:related}, prioritizing protocols that
stress generalization over saturated academic sets. The suite comprises
the thirteen benchmarks in Table~\ref{tab:benchmarks} and nineteen public
datasets; Table~\ref{tab:summary} lists every evaluation and headline
result. The suite is vision-only by design; audio
deepfakes are out of scope (\S\ref{sec:limitations}). For each benchmark
we fix the dataset version at download time, follow the authors' published
protocol, and record the license basis on which we use it. Results for the
Sumsub robustness benchmark are reported in \S\ref{sec:sumsub}; the
remaining benchmarks are run under the same single-snapshot, single-config
protocol.

\paragraph{GAS-Station.}
Alongside the public suite we release GAS-Station, an adversarial dataset
of AI-generated media produced by the SN34 competition itself. The image
split contains $50{,}399$ images from $78$ distinct generative
participants: a mix of named open models and adversarial miner
submissions with undisclosed pipelines, each sample carrying its prompt,
generation-week partition (a ten-week window, May--July 2026), and
provenance metadata (C2PA where available). A companion video split is
partitioned on the same weekly schedule. Both splits are available on
Hugging Face (\url{https://huggingface.co/datasets/gasstation/gs-images-v4},
\url{https://huggingface.co/datasets/gasstation/gs-videos-v4}). Because
the detector population co-evolves with this distribution, scoring a
snapshot on weeks concurrent with its training would amount to
evaluating on training-adjacent data; weeks generated \emph{after} a
snapshot's export are immune to this objection. The released window
postdates the evaluated April~15 export, and we use its most recent
weeks as a forward test in \S\ref{sec:temporal}.
Beyond that, GAS-Station's role is the opposite: to let third parties
train and test detectors against the adversarial frontier the mechanism
continuously generates.

\begin{table}[t]
\centering\small
\caption{Evaluation suite. All benchmarks
are evaluated with one dated BMF export and no per-benchmark tuning.
DeepfakeBench supplies the datasets and protocol for the face
cross-dataset rows rather than a result row of its own.}
\label{tab:benchmarks}
\setlength{\tabcolsep}{4pt}
\begin{tabular}{ll}
\toprule
Benchmark & Domain \\
\midrule
Sumsub~\cite{sumsub}              & In-the-wild robust. \\
Deepfake-Eval-2024~\cite{deepfakeeval2024} & In-the-wild \\
DeepfakeBench~\cite{deepfakebench} & Face cross-dataset \\
Community For.~\cite{communityforensics} & AI-gen image (21 gen.) \\
AIGCDetectBench~\cite{npr}        & AI-gen image (17 gen.) \\
GenImage~\cite{genimage}          & AI-gen image (legacy) \\
AIGIBench~\cite{aigibench}        & AI-gen robustness \\
WildRF~\cite{wildrf}              & In-the-wild \\
RAID~\cite{raid}                  & Adversarial \\
DF40~\cite{df40}                  & Face cross-dataset \\
GenVidBench~\cite{genvidbench}    & AI-gen video \\
GenVideo~\cite{genvideo}          & AI-gen video \\
AI-GenBench~\cite{aigenbench}     & AI-gen image (36 gen.) \\
\bottomrule
\end{tabular}
\end{table}

\subsection{Baselines}\label{sec:baselines}
All baseline numbers come from published results; we run only BMF through
our harness. For each comparison we cite the number reported by the
benchmark's authors or by the baseline's original paper at the matching
protocol, and we state the protocol next to every comparison so that any
residual mismatch in training data, input pipeline, or aggregation level
is explicit. Baselines are chosen per benchmark. For the face cross-dataset protocol
(\S\ref{sec:face-video}) we compare against the current frontier,
Effort~\cite{effort}; for the Sumsub robustness benchmark
(\S\ref{sec:sumsub}) we use the detectors its authors
report (Xception, MAT, M2TR, RECCE, CADDM, SBI). For AI-generated images
(\S\ref{sec:image}) the baselines are the Community Forensics detector
and the prior methods it reports
(CNNSpot~\cite{cnnspot}, UniFD~\cite{unifd},
RED140~\cite{red140}).

\subsection{Metrics}
We report ROC-AUC (primary), balanced accuracy, TPR at $1\%$ FPR, equal
error rate, and expected calibration error, with DeLong~\cite{delong}
$95\%$ confidence intervals for AUC. Because baselines are cited rather than re-run
(\S\ref{sec:baselines}), we compare published point values against our
confidence intervals rather than running paired tests against baselines.
Sumsub metrics are quoted in percent, following that benchmark's
convention; all other AUCs are decimals in $[0,1]$ (Sumsub values are
converted to decimals in cross-benchmark tables). In all tables, ``--''
marks an unreported or inapplicable value. We foreground TPR at low FPR
because, in production moderation, the cost
of false positives on authentic media is high, so the operationally
relevant quantity is recall at a fixed, low false-positive budget rather
than a threshold-free average; a detector with strong AUC can still be
unusable if its recall collapses once the threshold is set conservatively.
This is also the axis on which an independent audit of public detection
tools, which evaluated BMF among others, found low recall in
practice~\cite{publictools}, so we confront it directly rather than
reporting AUC and accuracy alone.

\section{Results}\label{sec:results}
BMF deploys three checkpoints: an image model, a general video model,
and a human-video model for face-centric deepfake video
(\S\ref{sec:arch}). We group results by the model under test: image
benchmarks (\S\ref{sec:results-image}) and video benchmarks
(\S\ref{sec:results-video}, which identifies the video checkpoint used
per benchmark), followed by a cross-cutting analysis of
adaptation over time (\S\ref{sec:temporal}). Table~\ref{tab:summary}
summarizes the headline results across the full suite.

\begin{table}[t]
\centering\small
\caption{Full summary of BMF results across all evaluated benchmarks
(headline ROC-AUC; per-benchmark sections give full metrics and baselines).
Face cross-dataset entries are scored without deliberately training on
any of the listed datasets (BMF is not FF++-trained); the face-video rows
are additionally audited for training near-duplicates
(\S\ref{sec:limitations}), and Celeb-DF values are
contamination-audited clean subsets
(Table~\ref{tab:face-video}); RAID
shows the clean-set AUC (adversarial curve in Table~\ref{tab:raid}).}
\label{tab:summary}
\setlength{\tabcolsep}{5pt}
\begin{tabular}{llc}
\toprule
Benchmark & Domain & AUC \\
\midrule
\multicolumn{3}{l}{\emph{Image (image checkpoint)}}\\
Sumsub (orig.\ / all cond.) & In-the-wild robust. & 0.936 / 0.872 \\
Deepfake-Eval-2024   & In-the-wild         & 0.915 \\
WildRF-Reddit        & In-the-wild         & 0.997 \\
WildRF-Twitter       & In-the-wild         & 0.998 \\
WildRF-Facebook      & In-the-wild         & 0.999 \\
Community Forensics  & AI-gen image        & 0.991 \\
AIGCDetectBench      & AI-gen image        & 0.999 \\
GenImage             & AI-gen image        & 0.9999 \\
AI-GenBench          & AI-gen image        & 0.995 \\
AIGIBench            & AI-gen robustness   & 0.998 \\
RAID (clean)         & Adversarial         & 0.969 \\
UADFV                & Face cross-dataset  & 0.970 \\
DF40                 & Face cross-dataset  & 0.950 \\
DFD                  & Face cross-dataset  & 0.932 \\
FaceForensics++      & Face cross-dataset  & 0.811 \\
DFDC-faces           & Face cross-dataset  & 0.674 \\
\midrule
\multicolumn{3}{l}{\emph{Video (general video checkpoint)}}\\
Deepfake-Eval-2024   & In-the-wild         & 0.822 \\
GenVidBench          & AI-gen video        & 0.918 \\
GenVideo-100K        & AI-gen video        & 0.924 \\
\midrule
\multicolumn{3}{l}{\emph{Video (human-video checkpoint)}}\\
DFDC                 & Face cross-dataset  & 0.947 \\
Celeb-DF v2          & Face cross-dataset  & 0.9985 \\
Celeb-DF v1          & Face cross-dataset  & 0.982 \\
Celeb-DF++           & Face cross-dataset  & 0.867 \\
\bottomrule
\end{tabular}
\end{table}

\subsection{Comparison to Best Published Results}\label{sec:vsbest}
Table~\ref{tab:vsbest} summarizes the strongest published reference point
available for benchmarks where a meaningful comparison can be made. The
table is not a claim that every protocol is identical: where the published
metric, split, or aggregation differs from ours, the mismatch is stated in
the caption and discussed in the corresponding subsection. On
Deepfake-Eval-2024,
BMF is at statistical parity with the best commercial detector on the
image track (the published $0.90$ lies inside our $95\%$ CI,
\S\ref{sec:analysis}) and is numerically above it on video ($0.79$ falls
just below our CI; \S\ref{sec:analysis}). Most strikingly, BMF's video model \emph{exceeds} the
FF++-trained specialist Effort on DFDC and Celeb-DF~v2 while scoring
contamination-audited evaluation subsets,
reversing the usual generalist
disadvantage (\S\ref{sec:face-video}). The saturated suites
(GenImage, AIGCDetectBench) sit at parity near
$1.0$, and Celeb-DF++, the hardest of the face-video set, is at
statistical parity with its specialist reference: the published $0.851$
lies inside BMF's $95\%$ CI ($0.867$ $[0.845, 0.889]$;
Table~\ref{tab:face-video}).

\begin{table}[t]
\centering\small
\caption{BMF vs.\ the best previously published or benchmark-reported
reference result where a meaningful comparison is available. Best-published values are the strongest
figures reported in the cited sources (for Deepfake-Eval-2024,
the benchmark authors' best anonymized commercial detector); metric is ROC-AUC except
where marked $\dagger$ (average precision). BMF uses one dated export with no per-benchmark
training. $^\ast$ parity: the
published value lies inside BMF's $95\%$ CI (\S\ref{sec:analysis}). Bold
entries are large-margin improvements over the published value under the
stated protocol. Sumsub row uses the original-condition,
generator-averaged protocol of the source paper; the all-conditions
pooled AUC is 0.872 (Table~\ref{tab:stats}).
Community Forensics reports the source detector's per-generator mAP and
BMF's pooled AP over our standard 21-generator evaluation panel
(Table~\ref{tab:commforensics}), so that row is indicative rather than a
strict metric match. For DF40, Effort's $0.940$ is the unweighted mean
of its eight cross-method subset AUCs ($0.9395$ before rounding), whereas
BMF's $0.950$ is pooled over evaluation frames; that row is therefore
indicative rather than a strict aggregation match. $^{v}$~scored by the human-video
checkpoint (Table~\ref{tab:face-video}); the image checkpoint's
frame-level DFDC score is far lower (Table~\ref{tab:face-cross}).}
\label{tab:vsbest}
\setlength{\tabcolsep}{3pt}
\begin{tabular}{llcc}
\toprule
Benchmark & Setting & Best pub. & BMF \\
\midrule
Deepfake-Eval (img)~\cite{deepfakeeval2024} & In-the-wild & 0.90 & 0.915$^\ast$ \\
Deepfake-Eval (vid)~\cite{deepfakeeval2024} & In-the-wild & 0.79 & 0.822 \\
Community Forensics~\cite{communityforensics} & OOD image & 0.987$^\dagger$ & 0.994$^\dagger$ \\
Sumsub (orig.)~\cite{sumsub} & In-the-wild & 0.857 & \textbf{0.936} \\
DF40~\cite{df40,effort} & Cross-method & 0.940 & 0.950 \\
Celeb-DF v2$^{v}$~\cite{celebdf} & Face x-data & 0.956 & \textbf{0.9985} \\
DFDC$^{v}$~\cite{dfdc} & Face x-data & 0.843 & \textbf{0.947} \\
\bottomrule
\end{tabular}
\end{table}

\subsection{Image Benchmarks}\label{sec:results-image}
These benchmarks exercise the image detector; face-deepfake datasets are
sourced from video but scored frame-level.

\subsubsection{In-the-Wild Robustness: Sumsub}\label{sec:sumsub}
Protocol: face-swap fakes (SimSwap~\cite{simswap} and Inswapper over LFW, CelebA-HQ,
FairFace) vs.\ the corresponding real sets; manipulations applied at
evaluation time with the authors' pipeline (JPEG $q{=}75$, downscale to
128px, GPEN enhancement~\cite{gpen}), matching the benchmark's protocol~\cite{sumsub}. The pooled
and per-generator results (Tables~\ref{tab:sumsub-pooled}--\ref{tab:sumsub-main})
are computed over $843{,}692$ images spanning the original and
GPEN-enhanced conditions; the robustness battery
(Table~\ref{tab:sumsub-robust}) breaks performance out by condition and
additionally scores the downscale-128 and JPEG-75 sets ($281{,}861$
images each), for $1{,}407{,}414$ scored images in total. We take the
original-condition AUC ($93.6$, Table~\ref{tab:sumsub-robust}) and the
all-conditions pooled AUC ($0.872$, Table~\ref{tab:stats}) as the
headline numbers; the two-condition pool of
Table~\ref{tab:sumsub-pooled} reflects a single scoring run and is
reported for completeness.

\begin{table}[t]
\centering\small
\caption{Pooled performance of our snapshot over the Sumsub original
and GPEN-enhanced conditions (843{,}692 samples). Threshold-free metrics (AUC, AP, EER)
summarize ranking quality; accuracy and balanced accuracy are reported
at a fixed $0.5$ decision threshold. The pool is fake-heavy, so average
precision is elevated by class prevalence; the pooled AUC over all four
conditions is $0.872$ (Table~\ref{tab:stats}).}
\label{tab:sumsub-pooled}
\begin{tabular}{lc}
\toprule
Metric & Value \\
\midrule
ROC-AUC            & 94.6 \\
Average precision  & 98.9 \\
EER                & 11.5 \\
Accuracy @ 0.5     & 67.8 \\
Balanced accuracy  & 80.5 \\
\bottomrule
\end{tabular}
\end{table}

\begin{table}[t]
\centering\small
\caption{ROC-AUC (\%) on the Sumsub benchmark (original images).
``Worst'' is the lowest of the three source datasets (LFW, CelebA-HQ,
FairFace); per-source values for our snapshot appear in
Table~\ref{tab:sumsub-persource}. Baseline numbers from the original
paper; BMF run under identical protocol.}
\label{tab:sumsub-main}
\setlength{\tabcolsep}{4pt}
\begin{tabular}{lcccc}
\toprule
 & \multicolumn{2}{c}{SimSwap} & \multicolumn{2}{c}{Inswapper} \\
\cmidrule(lr){2-3}\cmidrule(lr){4-5}
Model & Overall & Worst & Overall & Worst \\
\midrule
FF (Xception)   & 51.7 & 47.6 & 56.7 & 49.2 \\
MAT             & 79.7 & 49.0 & 80.3 & 49.8 \\
M2TR            & 55.3 & 54.9 & 53.9 & 53.1 \\
RECCE           & 56.7 & 46.9 & 56.1 & 51.5 \\
CADDM           & 78.2 & 75.2 & 59.8 & 56.6 \\
SBI             & 95.5 & 93.6 & 75.9 & 70.0 \\
\midrule
\textbf{BMF (ours)} & \textbf{98.6} & \textbf{98.2} & \textbf{88.7} & \textbf{77.1} \\
\bottomrule
\end{tabular}
\end{table}

\begin{table}[t]
\centering\small
\caption{Per-source ROC-AUC (\%) for our snapshot on the Sumsub
benchmark, original (unmanipulated) images.}
\label{tab:sumsub-persource}
\begin{tabular}{lcc}
\toprule
Source & SimSwap & Inswapper \\
\midrule
LFW       & 98.3 & 90.0 \\
CelebA-HQ & 98.2 & 98.9 \\
FairFace  & 99.3 & 77.1 \\
\bottomrule
\end{tabular}
\end{table}

\begin{table}[t]
\centering\small
\caption{Robustness: ROC-AUC under manipulation attacks (overall,
SimSwap/Inswapper averaged), over the full manipulation battery
($1{,}407{,}414$ scored images across the original and three manipulation
conditions). Baselines from the original paper's Table~2.}
\label{tab:sumsub-robust}
\setlength{\tabcolsep}{4pt}
\begin{tabular}{lcccc}
\toprule
Model & Original & Downscale & JPEG(75) & GPEN \\
\midrule
SBI (best prior) & 85.7 & 36.1 & 74.7 & 75.1 \\
CADDM            & 69.0 & 59.4 & 62.6 & 57.0 \\
MAT              & 80.0 & \textbf{82.8} & \textbf{89.2} & 79.3 \\
\midrule
\textbf{BMF (ours)} & \textbf{93.6} & 79.9 & 85.5 & \textbf{99.6} \\
\bottomrule
\end{tabular}
\end{table}

Two findings stand out. First, BMF's advantage is largest exactly where
prior detectors are weakest. On original images it improves on the
strongest self-blending baseline (SBI) by $+3.1$ AUC on SimSwap and
$+12.8$ on Inswapper (Table~\ref{tab:sumsub-main}), and the gap widens
under manipulation: SBI collapses to $36.1$ AUC under downscaling, while
BMF holds $85.5$ under JPEG and $99.6$ under GPEN
(Table~\ref{tab:sumsub-robust}).
We attribute this robustness primarily to full-frame scoring without a
face-crop stage (\S\ref{sec:input}), which sidesteps the crop-detector
failures several of these manipulations are designed to induce, and to
the degradation robustness of individual branches, chiefly ConvNeXt-L; a
per-branch analysis (\S\ref{sec:ablations}) shows that under heavy
degradation uniform fusion can dilute the strongest branch rather than
protect it. A DeLong $95\%$ confidence interval for the all-conditions
Sumsub pool is reported in \S\ref{sec:analysis} (Table~\ref{tab:stats}); at this sample size the
interval spans under half an AUC point, so the $3$--$13$ point margins
over SBI far exceed sampling uncertainty.

Second, BMF is not uniformly dominant. It trails MAT on downscaled inputs
($79.9$ vs $82.8$) and on JPEG ($85.5$ vs $89.2$), a weakness traceable to
Inswapper--FairFace, where down-128 scoring falls to 50.9\%: the
ensemble's high-resolution branches lose their edge when the input retains
little high-frequency detail. The Sumsub baselines, though the strongest
its authors evaluated, are 2019--2022 methods that predate the current
CLIP- and DINO-based frontier (\S\ref{sec:baselines}). We
therefore read Sumsub as a robustness stress test rather than a comparison
against the state of the art, and defer head-to-head evaluation against
current methods such as Effort to the face-deepfake cross-dataset protocol
in \S\ref{sec:face-video}.

\subsubsection{In-the-Wild: Deepfake-Eval-2024 (Image Track)}\label{sec:eval2024}
On the image track of Deepfake-Eval-2024 ($1{,}975$ in-the-wild images
circulated on social media in 2024, the setting of the steepest
documented in-the-wild AUC drops~\cite{deepfakeeval2024}), BMF attains
$0.915$ ROC-AUC with
$86.9\%$ balanced accuracy and $85.8\%$ accuracy at the default $0.5$
threshold (Table~\ref{tab:eval2024-img}), numerically
above the benchmark's best commercial detector ($0.90$; parity within our
$95\%$ CI, \S\ref{sec:analysis}). It far exceeds
the best open-source model (UniFD~\cite{unifd}, $0.56$, near chance).
Low-FPR operating points and calibration are reported in
\S\ref{sec:analysis} (Table~\ref{tab:stats}).

\begin{table}[t]
\centering\small
\caption{Deepfake-Eval-2024 image track ($1{,}975$ in-the-wild images).
Baselines from the benchmark authors~\cite{deepfakeeval2024} (their
Tables~5 and~7): UniFD~\cite{unifd} is their strongest off-the-shelf
open-source model; the commercial detector is their best-performing,
reported anonymously under contractual agreements (vendor pool: Hive,
Reality Defender, Pindrop, AI or Not, Hiya, Fraunhofer, Sensity AI).
Their strongest benchmark-finetuned model reaches only $0.73$ AUC.
BMF accuracy at the default $0.5$ threshold.}
\label{tab:eval2024-img}
\begin{tabular}{lccc}
\toprule
Method & ROC-AUC & Bal.\ Acc & Acc \\
\midrule
UniFD (open-source)~\cite{unifd}  & 0.56 & -- & -- \\
Best commercial~\cite{deepfakeeval2024}     & 0.90 & -- & 82 \\
\midrule
\textbf{BMF (ours)}      & \textbf{0.915} & 86.9 & 85.8 \\
\bottomrule
\end{tabular}
\end{table}

\subsubsection{Face-Deepfake Cross-Dataset (Image Model, Frame-Level)}\label{sec:face}
We evaluate BMF on five face-deepfake datasets under the standard
frame-level protocol: scoring sampled frames with the image checkpoint and
also aggregating to video-level by averaging each video's frame scores
(Table~\ref{tab:face-cross}). Because the fixed $0.5$ threshold is not recalibrated after
video-level averaging, video-level \emph{accuracy} understates performance
(e.g.\ FF++ accuracy falls to $31.1\%$ while balanced accuracy is
$54.5\%$); we therefore read AUC and balanced accuracy as the fair metrics.
Unlike the DeepfakeBench baselines, BMF is not trained on
FaceForensics++, so every dataset here, FF++ included, is
out-of-distribution for it.

\begin{table}[t]
\centering\small
\caption{The \emph{image checkpoint} on face-deepfake datasets,
frame-level / video-level (frame scores averaged per video). AUC and
balanced accuracy (\%). BMF is not FF++-trained, so all are cross-dataset
for it. DFD (Google/Jigsaw, shipped with FF++) is frame-level
only. The human-video specialist's much stronger results on the face
\emph{video} benchmarks (DFDC, Celeb-DF) appear in
Table~\ref{tab:face-video}; this table isolates what the image model
alone achieves on face content.}
\label{tab:face-cross}
\setlength{\tabcolsep}{4pt}
\begin{tabular}{lcccc}
\toprule
 & \multicolumn{2}{c}{AUC} & \multicolumn{2}{c}{Bal.\ Acc} \\
\cmidrule(lr){2-3}\cmidrule(lr){4-5}
Dataset & Frame & Video & Frame & Video \\
\midrule
UADFV~\cite{uadfv}          & 0.970 & 0.913 & 88.2 & 85.4 \\
DF40~\cite{df40}            & 0.950 & 0.938 & 87.0 & 80.5 \\
DFD~\cite{ffpp}             & 0.932 & --    & 84.7 & --   \\
FaceForensics++~\cite{ffpp} & 0.811 & 0.812 & 66.4 & 54.5 \\
DFDC (faces)~\cite{dfdc}    & 0.674 & 0.762 & 59.5 & 60.7 \\
\bottomrule
\end{tabular}
\end{table}

The spread is dataset-specific rather than a blanket face-swap weakness.
BMF generalizes strongly to UADFV ($0.970$ frame AUC), the diffusion- and
modern-pipeline forgeries of DF40 ($0.950$), and DFD ($0.932$), but is
markedly weaker on two classic face-swap corpora: FaceForensics++ ($0.811$)
and especially DFDC ($0.674$ frame, $0.762$ video). We read this honestly:
BMF excels on the synthesis-heavy, in-the-wild, and diffusion-era content
that dominates its training distribution, but the compression-heavy, older
face-swap artifacts of FF++/DFDC, compounded by the
full-frame-versus-face-crop protocol gap (\S\ref{sec:input}), cost it
against specialists trained directly on those corpora.
For context, Effort reports a $0.940$ unweighted mean AUC across DF40's
eight cross-method subsets ($0.9395$ before rounding)~\cite{effort}.
BMF's $0.950$ is instead pooled over evaluation frames, so the numerical
$0.010$ difference should not be interpreted as a matched-protocol win.

\subsubsection{AI-Generated Images}\label{sec:image}
We evaluate BMF on the Community Forensics benchmark, a many-generators
out-of-distribution protocol for AI-generated image
detection~\cite{communityforensics}. Our run covers $51{,}836$ images,
balanced between real and fake, spanning $21$ modern generators (BMF's
standard image-evaluation panel): recent diffusion models (FLUX, SD/LCM
variants, Imagen~3, Ideogram), the Midjourney and DALL-E families,
Kandinsky, Stable Cascade, and GANs (DFGAN, GALIP). Pooled metrics appear in
Table~\ref{tab:cf-pooled} and the comparison against prior detectors in
Table~\ref{tab:commforensics}.

\begin{table}[t]
\centering\small
\caption{BMF on Community Forensics: pooled metrics over $51{,}836$
balanced images across $21$ generators. AUC, AP, and EER are
threshold-free; balanced accuracy and MCC are evaluated at the fixed
$0.5$ decision threshold; Brier score is computed directly from the
predicted probabilities and does not depend on a decision threshold.}
\label{tab:cf-pooled}
\begin{tabular}{lc}
\toprule
Metric & Value \\
\midrule
ROC-AUC            & 0.991 \\
Average precision  & 0.994 \\
Balanced accuracy  & 96.6 \\
MCC                & 0.934 \\
Brier              & 0.027 \\
EER                & 2.9 \\
\bottomrule
\end{tabular}
\end{table}

\begin{table}[t]
\centering\small
\caption{AI-generated image detection on Community Forensics. Baseline and
reference numbers are from~\cite{communityforensics}, Table~2
(comprehensive column), reported as per-generator mean average precision
(mAP) and accuracy; the reference detector is a ViT-S/CLIP backbone
trained end-to-end. BMF is reported as pooled average precision and
balanced accuracy over the $21$-generator evaluation panel. Because the
published AP is averaged per generator while BMF's AP is pooled over
samples, the comparison is indicative of standing rather than a strict
metric match.}
\label{tab:commforensics}
\setlength{\tabcolsep}{8pt}
\begin{tabular}{lcc}
\toprule
Method & mAP/AP & Acc. \\
\midrule
CNNSpot (Wang \emph{et al.})~\cite{cnnspot}  & 0.537 & 51.3 \\
UniFD~\cite{unifd}                 & 0.592 & 54.0 \\
RED140~\cite{red140}               & 0.764 & 56.2 \\
GenImage-trained~\cite{genimage}   & 0.912 & 81.8 \\
Community Forensics                & 0.987 & 89.2 \\
\midrule
\textbf{BMF (ours)}                & \textbf{0.994} & \textbf{96.6} \\
\bottomrule
\end{tabular}
\end{table}

BMF's pooled AP and balanced accuracy are numerically above the strongest
published reference numbers (Table~\ref{tab:commforensics}; note the
metric caveat there).
The per-generator picture is sharply bimodal.
Eighteen of the $21$ generators are effectively solved (AUC $\ge 0.999$
and balanced accuracy $\ge 99\%$), spanning the full range of modern
diffusion, autoregressive, and GAN families, evidence that the snapshot
generalizes cleanly across generators it was not specifically tuned for.
Two are calibration-limited: DALL-E~2 (AUC $0.990$) and Firefly~Image~2
(AUC $0.997$) rank well but carry elevated Brier scores
(Table~\ref{tab:cf-hard}), the overconfident regime where a temperature
above $1$ would help. One is a genuine generalization gap: on Hourglass,
AUC falls to $0.856$ and balanced accuracy to $68.1\%$, a separation
failure rather than a calibration artifact, notably on the largest single
generator in the panel ($4{,}000$ images). We return to Hourglass in the
failure analysis (\S\ref{sec:failure}).

\begin{table}[t]
\centering\small
\caption{Hardest generators for BMF on Community Forensics (lowest AUC).}
\label{tab:cf-hard}
\begin{tabular}{lccc}
\toprule
Generator & AUC & Bal.\ Acc & Brier \\
\midrule
Firefly Image 2 & 0.997 & 94.5 & 0.040 \\
DALL-E 2        & 0.990 & 91.9 & 0.060 \\
Hourglass       & 0.856 & 68.1 & 0.234 \\
\bottomrule
\end{tabular}
\end{table}

Four further benchmarks corroborate the result
(Table~\ref{tab:aigc-extra}): BMF is near-saturated on the legacy suites,
under AIGIBench's degradation battery, and on the WildRF social-media
splits, with the only softness a $90.9\%$ balanced accuracy on the
heavily-recompressed Facebook split. The discriminating image results
therefore remain Community Forensics (Table~\ref{tab:commforensics}) and
Deepfake-Eval-2024.

\begin{table}[t]
\centering\small
\caption{BMF on further AI-generated image benchmarks. AIGIBench is the
mean over its $24$ generator subsets; WildRF is split by source platform.}
\label{tab:aigc-extra}
\setlength{\tabcolsep}{5pt}
\begin{tabular}{lccc}
\toprule
Benchmark & AUC & Bal.\ Acc & MCC \\
\midrule
GenImage~\cite{genimage}            & 0.9999 & 99.7 & 0.994 \\
AIGCDetectBench~\cite{npr}          & 0.999 & 99.4 & 0.988 \\
AIGIBench~\cite{aigibench}          & 0.998 & 97.3 & 0.948 \\
WildRF-Reddit~\cite{wildrf}         & 0.997 & 95.4 & 0.911 \\
WildRF-Twitter~\cite{wildrf}        & 0.998 & 94.8 & 0.886 \\
WildRF-Facebook~\cite{wildrf}       & 0.999 & 90.9 & 0.833 \\
\bottomrule
\end{tabular}
\end{table}

\paragraph{Adversarial robustness (RAID).}
We additionally stress BMF against transferable adversarial perturbations
using RAID~\cite{raid}, whose attacks are crafted against an ensemble of
seven detectors and transfer to unseen models. BMF degrades gracefully
rather than collapsing (Table~\ref{tab:raid}): from $0.969$ AUC on the
clean set it holds $0.821$ at $\epsilon{=}16$, falling to $0.695$ only
under the strongest $\epsilon{=}32$ budget. Large adversarial budgets do
erode the detector, an open problem for the whole field, but BMF retains
usable ranking quality across the practical perturbation range.

\begin{table}[t]
\centering\small
\caption{Adversarial robustness on RAID~\cite{raid}: ROC-AUC, balanced
accuracy, and MCC as the adversarial perturbation budget $\epsilon$
increases (transferable attacks).}
\label{tab:raid}
\begin{tabular}{lccc}
\toprule
Budget & AUC & Bal.\ Acc & MCC \\
\midrule
Clean ($\epsilon{=}0$) & 0.969 & 88.5 & 0.785 \\
$\epsilon{=}8$         & 0.894 & 76.4 & 0.578 \\
$\epsilon{=}16$        & 0.821 & 68.8 & 0.448 \\
$\epsilon{=}32$        & 0.695 & 60.5 & 0.275 \\
\bottomrule
\end{tabular}
\end{table}

\paragraph{AI-GenBench.}
We further evaluate on AI-GenBench~\cite{aigenbench}, scoring BMF zero-shot
across its $36$ generators ($14{,}400$ fakes, $12{,}000$ reals). BMF ranks
fakes near-perfectly ($0.995$ overall ROC-AUC, with per-generator AUC above
$0.96$ everywhere), but the fixed $0.5$ threshold exposes a recall gap:
overall fake recall is $86.3\%$, and on \emph{inpainting} and
partial-manipulation generators it falls sharply
($0.41$--$0.53$; Table~\ref{tab:aigenbench}) even
though their AUC stays near $0.98$. The pattern is diagnostic: when only a
local region is synthesized, the global $P(\text{fake})$ sits near the
decision boundary, so the model \emph{ranks} the fake correctly but the
default threshold misses it, an operating-point effect
(\S\ref{sec:analysis}). Fully synthetic modern
generators (DALL-E~3, FLUX, GANformer) are caught at $100\%$ recall.

\begin{table}[t]
\centering\small
\caption{AI-GenBench~\cite{aigenbench}: generators with the lowest recall at
the fixed $0.5$ threshold, with their (near-perfect) ROC-AUC. The gap
concentrates on inpainting / partial-manipulation generators.}
\label{tab:aigenbench}
\setlength{\tabcolsep}{5pt}
\begin{tabular}{lccl}
\toprule
Generator & Recall@0.5 & AUC & Type \\
\midrule
LaMa            & 0.41 & 0.980 & inpainting \\
Palette         & 0.51 & 0.985 & inpaint./diff. \\
MAT             & 0.52 & 0.987 & inpainting \\
SN-PatchGAN     & 0.53 & 0.992 & inpainting \\
DeepFloyd~IF    & 0.55 & 0.970 & diffusion \\
CycleGAN        & 0.67 & 0.992 & GAN \\
ADM             & 0.69 & 0.964 & diffusion \\
\midrule
DALL-E~3 / FLUX & 1.00 & --    & fully synth. \\
\bottomrule
\end{tabular}
\end{table}

\subsection{Video Benchmarks}\label{sec:results-video}
These benchmarks exercise the two video checkpoints
(\S\ref{sec:video-arch}): the general video model scores the
in-the-wild and AI-generated-video benchmarks
(\S\ref{sec:eval2024-vid}--\ref{sec:video}), and the human-video
specialist scores the face-deepfake benchmarks
(\S\ref{sec:face-video}).

\subsubsection{In-the-Wild: Deepfake-Eval-2024 (Video Track)}\label{sec:eval2024-vid}
On the video track ($814$ in-the-wild clips, $385$ fake / $429$ real), the
BMF general video checkpoint reaches $0.822$ ROC-AUC with $73.0\%$ accuracy and
$72.2\%$ balanced accuracy (Table~\ref{tab:eval2024-vid}). This is a solid
result in the regime where the benchmark's authors report the steepest
drop: it exceeds the benchmark's best commercial detector
($0.79$ AUC), far exceeds the best off-the-shelf open-source model
(GenConViT, $0.63$), and matches, without in-domain training, the authors' own
GenConViT finetuned on the benchmark's training split ($0.82$, inside our
$95\%$ CI). It trails our own image track ($0.915$), consistent with
video being the harder and less mature side of the system. At the default threshold the operating point is
conservative: BMF correctly flags $370/429$ real clips ($86.2\%$
specificity) but catches only $224/385$ fakes ($58.2\%$ recall), so the
fixed threshold trades recall for precision. We report this directly; the
calibration analysis (\S\ref{sec:analysis}) characterizes the low-FPR
operating points that govern this recall/precision trade-off. Taken with
the image track, this gives the cross-modal in-the-wild result that prior
single-modality evaluations lack.

\begin{table}[t]
\centering\small
\caption{Deepfake-Eval-2024 video track ($814$ in-the-wild clips).
Baselines from the benchmark authors~\cite{deepfakeeval2024}: GenConViT
is their strongest off-the-shelf open-source model; the commercial
detector is their best-performing, reported anonymously under
contractual agreements. Their GenConViT finetuned on the benchmark's
training split reaches $0.82$ AUC, which zero-shot BMF matches without
any in-domain training. BMF accuracy at the default $0.5$ threshold.}
\label{tab:eval2024-vid}
\begin{tabular}{lccc}
\toprule
Method & ROC-AUC & Bal.\ Acc & Acc \\
\midrule
GenConViT (open-source)~\cite{deepfakeeval2024}  & 0.63 & -- & -- \\
Best commercial~\cite{deepfakeeval2024}     & 0.79 & -- & 78 \\
\midrule
\textbf{BMF (ours)}      & \textbf{0.822} & 72.2 & 73.0 \\
\bottomrule
\end{tabular}
\end{table}

\subsubsection{AI-Generated Video}\label{sec:video}
On GenVidBench~\cite{genvidbench}, a benchmark of AI-generated video from
recent text-to-video systems (Sora, Kling, and others), the BMF video
checkpoint attains $0.918$ ROC-AUC with $84.0\%$ balanced accuracy over
a balanced $1{,}999$-clip subset of the benchmark ($1{,}000$ fake,
$999$ real), at a near-symmetric operating
point: fake recall $83.9\%$, real specificity $84.2\%$
(Table~\ref{tab:genvidbench}). This mirrors the image-side performance and
indicates that the separately trained video model transfers to
commercial-grade generated video without per-benchmark tuning. For
reference, the GenVidBench authors' strongest supervised baselines reach
$79.9\%$ Top-1 accuracy (MViT-V2, cross-generator split) and $85.5\%$
(DeMamba, full set)~\cite{genvidbench,genvideo}, against BMF's $84.0\%$
balanced accuracy. This is an indicative comparison rather than a strict
head-to-head: those baselines are supervised on GenVidBench's training
generators and report accuracy, whereas BMF is evaluated zero-shot.

On the larger GenVideo benchmark~\cite{genvideo} (the DeMamba
million-scale corpus), evaluated on a $100$K-clip subset, the video model
reaches $0.924$ ROC-AUC with $78.6\%$ accuracy, consistent with the
GenVidBench result (Table~\ref{tab:genvidbench}).

\begin{table}[t]
\centering\small
\caption{BMF (general video checkpoint) on AI-generated video benchmarks.
Balanced accuracy at the default $0.5$ threshold.}
\label{tab:genvidbench}
\begin{tabular}{lccc}
\toprule
Benchmark & AUC & Bal.\ Acc & MCC \\
\midrule
GenVidBench~\cite{genvidbench}  & 0.918 & 84.0 & 0.681 \\
GenVideo-100K~\cite{genvideo}   & 0.924 & 78.6 & 0.593 \\
\bottomrule
\end{tabular}
\end{table}

\subsubsection{Face-Deepfake Cross-Dataset (Human-Video Model)}\label{sec:face-video}
The human-video checkpoint is the face-cropped, V-JEPA-augmented
specialist of \S\ref{sec:video-arch}; the general video model scores the
AI-generated-video and in-the-wild benchmarks above. For the face-video
benchmarks, we score audit-verified evaluation subsets: a frame-level
near-duplicate audit checks each benchmark against the training corpora
of all three models (image, general video, and human video;
\S\ref{sec:limitations}), and clips flagged as training near-duplicates
are excluded.

On these audited subsets, BMF reaches $0.947$ AUC on DFDC ($95\%$ CI
$[0.941, 0.953]$), compared with
Effort's published $0.843$ under its canonical protocol. On Celeb-DF~v2,
BMF scores $0.9985$ $[0.9957, 1.000]$ ($n{=}1{,}010$), above Effort's
$0.956$, and on
Celeb-DF~v1 it scores $0.982$ $[0.974, 0.990]$ ($n{=}1{,}010$).
Celeb-DF++, the 2025
successor spanning $22$ recent face-swap, reenactment, and talking-face
pipelines, remains the hardest face-video benchmark in this group: BMF
reaches $0.867$ $[0.845, 0.889]$ on the clean subset, and Effort's
published $0.851$ lies inside this interval, so we report parity with
the specialist rather than a win. The human-video checkpoint also far
exceeds the image model's frame-aggregated scores on the same face video
(DFDC $0.947$ vs $0.762$, Table~\ref{tab:face-cross}), indicating that
face-swap \emph{video} benefits from the temporal, face-cropped
specialist.

\begin{table}[t]
\centering\small
\caption{BMF human-video checkpoint (never trained on FF++ or on any of
these benchmarks' training splits) vs.\ the
FF++-trained specialist Effort~\cite{effort} on face-deepfake cross-dataset
benchmarks (video-level ROC-AUC). BMF is scored exclusively on
audit-verified subsets: $\approx$1K clips per dataset containing no clip
our contamination audit flagged as a training near-duplicate
(\S\ref{sec:limitations}); $n$ in parentheses (reals/fakes). Celeb-DF
rows are clean-subset results after the audit. Effort's
figures are its published canonical-protocol results, so comparisons are
indicative rather than split-matched;
its Celeb-DF~v1 figure is not reported. BMF CIs are DeLong $95\%$
intervals computed from per-clip scores.
\textbf{Bold} = BMF exceeds the specialist (published value falls below
BMF's CI); $^\ast$ = statistical parity (published value lies inside the
CI).}
\label{tab:face-video}
\setlength{\tabcolsep}{3.5pt}
\begin{tabular}{lccc}
\toprule
Dataset & Effort & BMF ($n$) & $95\%$ CI \\
\midrule
DFDC~\cite{dfdc}             & 0.843 & \textbf{0.947} (535/476) & .941--.953 \\
Celeb-DF v2~\cite{celebdf}   & 0.956 & \textbf{0.9985} (555/455) & .9957--1.00 \\
Celeb-DF v1~\cite{celebdf}   & --    & 0.982 (622/388) & .974--.990 \\
Celeb-DF++~\cite{celebdfpp}  & 0.851 & 0.867$^\ast$ (439/661) & .845--.889 \\
\bottomrule
\end{tabular}
\end{table}

\subsection{Adaptation Over Time}\label{sec:temporal}
\begin{figure}[t]
\centering
\begin{tikzpicture}
\begin{axis}[
  width=0.92\linewidth, height=4.6cm,
  xlabel={Snapshot export date (2025--2026)},
  ylabel={ROC-AUC},
  xmin=0.8, xmax=6.9, ymin=0.83, ymax=0.95,
  xtick={1,2,3,4,5,6},
  xticklabels={Nov,Dec,Jan,Feb,Mar,Apr},
  ytick={0.84,0.86,0.88,0.90,0.92,0.94},
  legend pos=south east, legend style={font=\scriptsize},
  grid=major, grid style={dotted,gray!40},
  tick label style={font=\scriptsize}, label style={font=\scriptsize},
]
\addplot[black, thick, mark=*, mark size=2pt]
  coordinates {(1.23,0.842) (3.35,0.891) (6.50,0.902)};
\addlegendentry{Image model}
\addplot[black!55, thick, densely dashed, mark=triangle*, mark size=2pt]
  coordinates {(1.23,0.864) (3.94,0.904) (6.50,0.936)};
\addlegendentry{Video model}
\end{axis}
\end{tikzpicture}
\caption{Adaptation over time, plotted from Table~\ref{tab:temporal}:
AUC on media from generators absent from the static baseline's training
improves with snapshot recency on both tracks.}
\label{fig:temporal}
\end{figure}
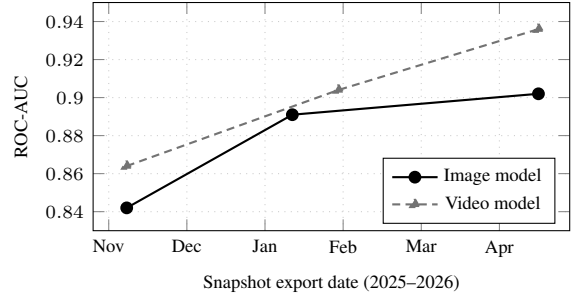

To test the central claim directly, we evaluate successive dated
snapshots of both models on fixed test sets of media from recent
generators absent from the static baseline's training, with matched reals
(Table~\ref{tab:temporal}, Figure~\ref{fig:temporal}). On the image
track ($25{,}099$ images), the static baseline (exported November~7,
2025) scores $0.842$ pooled AUC, and successive exports improve with
snapshot recency: $0.842 \to 0.891 \to 0.902$, a
gain of $6.0$ AUC points for the current snapshot over the static
baseline.
The video track (general video model) shows the same ordering on a test
set including Sora~2, Veo~3.1, Kling~2.6, and Wan~2.6:
three successive snapshots improve $0.864 \to 0.904 \to 0.936$,
a gain of $7.2$ points. Every refresh, on both tracks, improves AUC on
this fixed test set.
This study is a \emph{back-test}: earlier dated snapshots are evaluated on one common,
fixed test set. The test generators are absent from the training data
of the static baseline only; the later snapshots have seen media from
these generator families in training (though not the test items
themselves), by design, since incorporating new generators is
precisely what the mechanism does. The study therefore measures how
quickly the mechanism closes the gap a frozen model accumulates, not
zero-shot generalization by the refreshed snapshots (the static
baseline's score is the zero-shot number: it degrades gracefully rather
than collapsing on generator families it has never seen). Successive
snapshots build on successive competition rounds
(\S\ref{sec:incentive}) and may differ in architecture as well as
training data; each dated export, however, applies a roughly constant
in-house fine-tuning procedure to the then-current winning basis and
challenge distribution (\S\ref{sec:ensemble}), so the trajectory
reflects what the mechanism supplied rather than changes in our
post-processing. The study accordingly measures the mechanism
end-to-end rather than isolating data freshness
(\S\ref{sec:ablations}). That is the claim the deployment model rests
on: the mechanism, not any individual frozen export, is what keeps the
deployed detector current.

\paragraph{Forward evaluation on the live challenge stream.}
The back-test above holds the models fixed against curated sets of recent
generators; as a complement, we score the frozen image and video snapshots on
the live GAS-Station stream itself, using the most recent weeks available
at evaluation time (ISO weeks 2026-W26 through 2026-W28; $16{,}068$ verified
miner-generated images and $17{,}039$ videos), all generated more than two
months after the April~15 exports. No sample in this window
existed when the snapshots were trained, and week partitions are timestamped
in the public dataset, so the no-leakage guarantee is third-party auditable.
The image snapshot flags $92.3\%$ of these fakes at the fixed $0.5$ threshold
(Table~\ref{tab:gasbench-forward}), with high confidence (median
$P(\text{fake})$ of $0.98$), and recall is stable across the three weeks
($92.9\% \to 92.0\% \to 91.7\%$). The video snapshot flags $80.0\%$ of
post-export clips at the same threshold, with recall stable to slightly
rising across the window ($78.9\% \to 80.5\% \to 81.4\%$); the video stream,
dominated by frontier commercial generators, is the harder of the two, and
per-clip confidence is correspondingly lower (median $P(\text{fake})$ of
$0.82$). GAS-Station contains no authentic
media, so we report threshold metrics rather than AUC; the
corresponding real-side operating points are those of
\S\ref{sec:analysis}.

\begin{table}[t]
\centering\small
\caption{Forward evaluation of the frozen production snapshots on GAS-Station
weeks generated after their training cutoff (fake recall at the fixed $0.5$
threshold; the stream contains only synthetic media). W28 is partial
(through July 8, 2026).}
\label{tab:gasbench-forward}
\begin{tabular}{lrrr}
\toprule
Week (2026) & $n$ & Recall@$0.5$ & mean $P(\text{fake})$ \\
\midrule
\multicolumn{4}{l}{\emph{Image snapshot (Apr 15)}} \\
W26    & $6{,}751$  & 0.929 & 0.894 \\
W27    & $7{,}234$  & 0.920 & 0.877 \\
W28    & $2{,}083$  & 0.917 & 0.879 \\
Pooled & $16{,}068$ & 0.923 & 0.884 \\
\midrule
\multicolumn{4}{l}{\emph{Video snapshot (Apr 15)}} \\
W26    & $7{,}194$  & 0.789 & 0.710 \\
W27    & $6{,}924$  & 0.805 & 0.719 \\
W28    & $2{,}921$  & 0.814 & 0.729 \\
Pooled & $17{,}039$ & 0.800 & 0.717 \\
\bottomrule
\end{tabular}
\end{table}

\begin{table}[t]
\centering\small
\caption{Adaptation over time: ROC-AUC of successive dated snapshots,
back-tested on fixed test sets of recent-generator media. The test
generators are absent from the static baseline's training data; later
snapshots have seen media from these families (\S\ref{sec:temporal}).
Dates are 2026 except the static baseline (November~7, 2025).}
\label{tab:temporal}
\begin{tabular}{lc}
\toprule
Snapshot & AUC \\
\midrule
\multicolumn{2}{l}{\emph{Image track}} \\
Nov 7 (static)       & 0.842 \\
Jan 11               & 0.891 \\
BMF (Apr 15)         & \textbf{0.902} \\
\midrule
\multicolumn{2}{l}{\emph{Video track}} \\
Nov 7 (static)       & 0.864 \\
Jan 29               & 0.904 \\
BMF (Apr 15)         & \textbf{0.936} \\
\bottomrule
\end{tabular}
\end{table}

\section{Analysis}\label{sec:analysis}
\subsection{Failure Modes}\label{sec:failure}
The clearest failure is Hourglass on the Community Forensics panel (AUC
$0.856$). Grouping the panel by generator architecture is revealing:
Hourglass is the Hourglass Diffusion Transformer
(HDiT)~\cite{hdit}, which synthesizes images \emph{directly in pixel space}
rather than decoding from a latent autoencoder; the second-hardest
generator, DALL-E~2 (AUC $0.990$), is likewise a pixel-space cascaded
diffusion model (unCLIP) with no VAE stage. Every latent-diffusion
generator in the panel (FLUX, the Stable-Diffusion/LCM variants,
Kandinsky, Stable Cascade, Ideogram, the Midjourney family, and
DALL-E~3) is solved at AUC $\ge 0.999$. This pattern is consistent with
BMF keying, in part, on the characteristic upsampling fingerprint left by
latent-diffusion VAE decoders: generators lacking that decoder stage emit
low-level statistics further from the bulk of the training distribution.

We caution that the rule is not clean. The two GANs in the panel (DFGAN,
GALIP) also lack a VAE yet are detected near-perfectly, presumably through
the distinct and well-studied fingerprints of GAN upsamplers; and
Imagen~3, whose lineage is pixel-space cascaded diffusion, is also solved.
The operative factor is therefore narrower than ``pixel versus latent'':
likely the specific upsampler signature combined with each family's
representation in recent training rounds. We read Hourglass as
a probable new-family generalization gap: its AUC of $0.856$ indicates
imperfect class separation, while its elevated Brier score of $0.234$
indicates poor probabilistic accuracy arising from misclassification,
miscalibration, or both. This is the kind of gap the incentive mechanism
(\S\ref{sec:incentive}) is designed to close once such generators enter the
challenge distribution. Confirming this account, by auditing pixel-space
generator coverage in recent training rounds and testing whether targeted
fine-tuning recovers the Hourglass AUC, is left to future work.

\subsection{Calibration and Operating Points}
Threshold-free AUC can hide two production-relevant weaknesses: poor recall
at the low false-positive rates moderation requires, and miscalibrated
probabilities. We therefore report, for the primary benchmarks, the ROC-AUC
with a DeLong $95\%$ confidence interval, the true-positive rate at $1\%$
FPR, and the expected calibration error (Table~\ref{tab:stats}). Confidence
intervals are computed from per-sample scores; paired DeLong tests on
shared samples are significant for every BMF-internal comparison we run
(e.g., ensemble vs.\ single branches, \S\ref{sec:ablations}).

\begin{table}[t]
\centering\small
\caption{Statistical detail for the primary benchmarks: ROC-AUC with DeLong $95\%$
confidence interval, TPR at $1\%$ FPR, and expected calibration error
(ECE). DFE = Deepfake-Eval-2024. RAID rows trace the adversarial-budget
degradation.}
\label{tab:stats}
\setlength{\tabcolsep}{3pt}
\begin{tabular}{lccc}
\toprule
Benchmark & AUC (95\% CI) & TPR@1\% & ECE \\
\midrule
GenImage         & 0.9999 (.9997--.9999) & 0.999 & 0.029 \\
AIGCDetect       & 0.999 (.999--1.00)   & 0.979 & 0.028 \\
Community For.   & 0.991 (.991--.992)   & 0.958 & 0.034 \\
WildRF-R         & 0.997 (.994--.999)   & 0.972 & 0.083 \\
WildRF-T         & 0.998 (.995--.999)   & 0.957 & 0.078 \\
WildRF-F         & 0.999 (.997--1.00)   & 0.988 & 0.134 \\
DFE-image        & 0.915 (.893--.932)   & 0.580 & 0.070 \\
DFE-video        & 0.822 (.791--.848)   & 0.236 & 0.083 \\
DFD (face)       & 0.932 (.927--.937)   & 0.604 & 0.070 \\
Sumsub, all cond.\ & 0.872 (.870--.873) & 0.532 & 0.479 \\
RAID, clean      & 0.969 (.967--.973)   & 0.710 & 0.092 \\
RAID, $\epsilon{=}8$  & 0.894 (.887--.900) & 0.436 & 0.145 \\
RAID, $\epsilon{=}16$ & 0.821 (.811--.827) & 0.261 & 0.186 \\
RAID, $\epsilon{=}32$ & 0.695 (.682--.703) & 0.114 & 0.217 \\
\bottomrule
\end{tabular}
\end{table}

Three points follow. First, on the close Deepfake-Eval-2024 image margin,
the best published $0.90$ falls \emph{inside} our $95\%$ CI
$[0.893, 0.932]$, so we report image-track parity rather than a
significant win; on video, the published $0.79$ falls just below our CI
$[0.791, 0.848]$; because the published value is reported to only two
decimals, we read this as at most a marginal improvement rather than a
decisive one. Second, the low-FPR
operating points are honest about the recall critique
of~\cite{publictools}: at $1\%$ FPR the image detector recovers $58\%$ of
fakes but the video detector only $24\%$, confirming that in-the-wild
video recall at a strict threshold remains an open weakness even where AUC
is strong. Third, calibration is strongly dataset-dependent. It is
reasonable on most AIGC and in-the-wild sets (ECE $0.03$--$0.09$), but
breaks down in two regimes: the heavily fake-imbalanced,
manipulation-degraded Sumsub battery, where the model is badly overconfident (ECE
$0.479$; pooled AUC over \emph{all} conditions is $0.872$, versus $0.946$
over the original and GPEN conditions, \S\ref{sec:sumsub}); and under recompression or adversarial
perturbation (WildRF-Facebook $0.134$; RAID up to $0.217$ at $\epsilon{=}32$).
The two Deepfake-Eval tracks tilt the opposite way: the model is mildly
\emph{under}confident (predicting $\approx0.65$
where $\approx0.87$ of samples are fake), the benign direction for a
moderation operating point.

\subsection{Ablations}\label{sec:ablations}
We ablate the ensemble on a heterogeneous held-out pool of roughly
$97$K images drawn from three complementary sources: organic in-product
submissions (real uploads and flagged AI images from live traffic, with
labels manually reviewed for correctness), date-binned in-the-wild web
scrapes of real and AI imagery including very recent generators, and
curated public benchmarks spanning the major generative families (GAN,
diffusion, face-swap/reenactment) paired with diverse real-image
domains (faces, animals, street scenes, documents, art). The pool
covers $31$ distinct fake domains, and because each source stresses a
different generator or real distribution, per-domain scoring exposes
where individual experts collapse (Table~\ref{tab:ablation}).

Pooled, every branch trails the full ensemble: BMF reaches $0.934$ AUC,
$1.2$ points above the strongest single branch (CLIP ViT-L, $0.922$),
and the gain is statistically unambiguous (paired DeLong on shared
samples, $z{=}39$, $p\approx 0$). The stronger finding, however, is in
the worst-case column. Every individual branch fails catastrophically
on some fake domain, and each on a \emph{different} one: DINOv3 falls
to $0.211$ on FaceForensics++ face-swaps (worse than chance, i.e.,
blind to an entire deepfake class) where ConvNeXt-L and EVA-L score
$0.95$; ConvNeXt-L falls to $0.586$ on digi2real, which DINOv3 covers
at $0.992$; CLIP falls to $0.802$ on flickr-SD3, which EVA-L covers at
$0.956$. BMF is the only configuration that stays at or above $0.85$ on
all $31$ domains. On $23$ of the $31$ domains some individual branch
beats the ensemble, but the ensemble is never the
catastrophic one: it tracks the per-domain oracle without the tail
risk. This worst-case elimination, rather than the pooled margin, is
what heterogeneous fusion buys, and it is the property that matters
when serving traffic of unknown provenance.

The picture shifts under degradation. Re-running the per-branch
analysis on the hardest slice of the Sumsub battery (the LFW sources,
$119$K images), ConvNeXt-L is the clear robustness leader ($0.83$ AUC
under downscale-128, $0.82$ under JPEG-75), while DINOv3 is near-random
on face-swap content across all conditions and EVA-L collapses under
JPEG ($0.61$); uniform fusion consequently scores \emph{below}
ConvNeXt-L alone on degraded inputs ($0.73$/$0.79$).
Quality-conditioned branch routing improves every per-condition AUC but
breaks the global score ranking, so it does not help at a single
production threshold. The same conclusion holds in both analyses: no
single branch is strongest in every regime (CLIP ViT-L leads on clean
wild content but falls to $0.70$ under downscaling, where ConvNeXt-L
leads), so any single-branch deployment would carry a
distribution-specific failure mode. Closing the remaining gap to MAT on
degraded face-swap inputs (\S\ref{sec:sumsub}) likely requires
training-time degradation augmentation rather than any inference-time
fusion or routing scheme.

An ablation isolating training-data freshness (the same model retrained without the
most recent months of data) is left to future work; the temporal study
(\S\ref{sec:temporal}) provides the system-level version of that
evidence.

\begin{table}[t]
\centering\small
\caption{Branch ablation on a heterogeneous $\approx$97K-image holdout
(organic production traffic with manually reviewed labels, date-binned
web scrapes including recent generators, and curated public benchmarks;
$31$ fake domains). Worst = lowest per-domain AUC; the final column
counts fake domains below $0.85$ AUC. Branch labels follow
Table~\ref{tab:branches}.}
\label{tab:ablation}
\setlength{\tabcolsep}{3pt}
\begin{tabular}{lccc}
\toprule
Branch & AUC & Worst domain & \#$<$0.85 \\
\midrule
ConvNeXt-L (A1) & 0.887 & 0.586 (digi2real)  & 10 \\
EVA-L (A2)      & 0.898 & 0.778 (in-the-wild) & 8 \\
CLIP ViT-L (B)  & 0.922 & 0.802 (flickr-SD3) & 2 \\
DINOv3 (C)      & 0.889 & 0.211 (FF++ swaps) & 7 \\
\midrule
BMF             & \textbf{0.934} & \textbf{0.851} & \textbf{0} \\
\bottomrule
\end{tabular}
\end{table}

\section{Limitations}\label{sec:limitations}
Several caveats bound our claims. First, we evaluate a dated BMF export,
not the live system. The adaptation argument rests on the export's currency
at release and the temporal study in \S\ref{sec:temporal}; the deployed
system continues to evolve beyond the reported figures. Second,
our fixed image and general-video protocols score full frames with no face-crop
stage (\S\ref{sec:input}); on benchmarks whose baselines preprocess with
face detection this is a deliberate handicap we accept rather than tune
around, and some per-source weaknesses (notably low-resolution FairFace)
likely reflect it. Third, some benchmarks are excluded for licensing or
access reasons, notably Chameleon, which is pending permission; our
coverage of hard community images is therefore incomplete. Fourth, audio
deepfakes are out of scope: BMF dispatches audio to a separate model not
evaluated here, and the benchmark suite is vision-only. Fifth, because BMF trains on a large
continuously scraped corpus, we proactively audited the face-video
cross-dataset benchmarks (Table~\ref{tab:face-video}) for training
near-duplicates (frame-level
perceptual hashing against all three corpora), rescoring each dataset on
a $\approx$1K-clip flag-free subset. DFDC was fully clean; for the
Celeb-DF family, we report only the audit-verified subsets
(Table~\ref{tab:face-video}). Rescoring with the flagged media included
shifts AUC by at most $0.017$, so the flagged media were not driving
performance and are being purged from subsequent training rounds. The saturated public
image suites (GenImage, AIGCDetectBench, WildRF) were likewise verified
free of training overlap. The audit covers media-level duplication;
person-level identity overlap (the same public
figures appearing in both training and evaluation footage) is inherent
to celebrity-sourced benchmarks and is not excluded. Finally, the public benchmarks we use are predominantly
face- and photo-centric, which does not match the full modality mix of our
production traffic; our numbers should be read as benchmark performance,
not as a direct estimate of in-deployment accuracy.

\section{Conclusion}
The gap between benchmark and real-world deepfake detection is structural:
a detector frozen against a fixed corpus decays as the generative frontier
moves past it. Closing this gap requires a \emph{process}, not only an
architecture. We presented
BitMind Forensics (BMF), a detector built on the outputs of an open,
adversarial incentive mechanism that continually refreshes its
architecture basis and training distribution. Evaluated as a single dated
export with no per-benchmark tuning, BMF
reaches 0.936 AUC on original images and 0.872 pooled AUC over the full
manipulation battery of the Sumsub in-the-wild robustness benchmark. It
exceeds the strongest prior baseline on each generator and remains robust
where most prior detectors degrade ($0.855$ AUC under JPEG; $0.996$ after
GPEN enhancement). It reaches $0.915$ AUC on the
Deepfake-Eval-2024 image track, $0.991$ AUC on a 21-generator AI-image
panel, $0.918$ AUC on AI-generated video (GenVidBench), and $0.822$ AUC on
the Deepfake-Eval-2024 video track. Despite never training on
FaceForensics++, its video model exceeds the FF++-trained specialist on
DFDC ($0.947$) and Celeb-DF~v2 ($0.9985$, contamination-audited clean
subset), with statistical parity on Celeb-DF++. A temporal study
(\S\ref{sec:temporal}) further shows that successive exports improve on
media from generators absent from a static baseline's training, on both the image
($0.842 \to 0.902$) and video ($0.864 \to 0.936$) tracks.

The evidence supports the central claim: because the system's training
distribution tracks the generative frontier endogenously, even a static
export benefits from more current training data than statically trained
detectors. Our evaluation harness is public and at
publication the production API serves the exact evaluated snapshot, so
the protocol and deployed system can be independently checked. Continuous,
incentive-driven data refresh is a general strategy for non-stationary
detection problems, of which deepfake detection is a pressing example.

\section*{Reproducibility Statement}
The evaluation harness (dataset loaders, manipulation pipelines, and
scoring code) is the publicly available gasbench framework,
\url{https://github.com/BitMind-AI/gasbench}, and the incentive mechanism
that produces the evaluated detector is open-sourced at
\url{https://github.com/BitMind-AI/bitmind-subnet}. Dataset versions are
fixed at download time and recorded in our protocol (\S\ref{sec:protocol}).
We do not run any third-party detectors (baselines are cited from
publications, \S\ref{sec:baselines}), so no third-party checkpoints or API
versions are involved. Model weights and per-sample outputs are not public.
At publication, the production API serves the evaluated snapshot, so any
benchmark can be independently re-scored end-to-end using the
public harness. Because the deployed system continues to evolve
(\S\ref{sec:limitations}), re-scores after a subsequent export verify the
then-current system rather than the archival figures reported here.

\clearpage
{\small
\bibliographystyle{ieeetr}
\bibliography{references}

@inproceedings{deepfakeeval2024,
  author    = {Chandra, Nuria Alina and others},
  title     = {{Deepfake-Eval-2024}: A Multi-Modal In-the-Wild Benchmark of {Deepfakes} Circulated in 2024},
  booktitle = {CVPR Workshops},
  year      = {2026},
  note      = {arXiv:2503.02857}
}

@inproceedings{sumsub,
  author    = {Pirogov, Viacheslav and Artemev, Maksim},
  title     = {Evaluating {Deepfake} Detectors in the Wild},
  booktitle = {ICML DataWorld Workshop},
  year      = {2025},
  note      = {arXiv:2507.21905}
}

@inproceedings{df40,
  author    = {Yan, Zhiyuan and others},
  title     = {{DF40}: Toward Next-Generation {Deepfake} Detection},
  booktitle = {NeurIPS Datasets \& Benchmarks},
  year      = {2024},
  note      = {arXiv:2406.13495}
}

@inproceedings{deepfakebench,
  author    = {Yan, Zhiyuan and others},
  title     = {{DeepfakeBench}: A Comprehensive Benchmark of {Deepfake} Detection},
  booktitle = {NeurIPS Datasets \& Benchmarks},
  year      = {2023},
  note      = {arXiv:2307.01426}
}

@inproceedings{uadfv,
  author    = {Yang, Xin and Li, Yuezun and Lyu, Siwei},
  title     = {Exposing Deep Fakes Using Inconsistent Head Poses},
  booktitle = {ICASSP},
  year      = {2019},
  note      = {arXiv:1811.00656}
}

@misc{dfdc,
  author    = {Dolhansky, Brian and others},
  title     = {The {DeepFake} Detection Challenge ({DFDC}) Dataset},
  year      = {2020},
  note      = {arXiv:2006.07397}
}

@inproceedings{celebdf,
  author    = {Li, Yuezun and Yang, Xin and Sun, Pu and Qi, Honggang and Lyu, Siwei},
  title     = {{Celeb-DF}: A Large-scale Challenging Dataset for {DeepFake} Forensics},
  booktitle = {CVPR},
  year      = {2020},
  note      = {arXiv:1909.12962}
}

@misc{celebdfpp,
  author    = {Li, Yuezun and others},
  title     = {{Celeb-DF++}: A Large-scale Challenging Video {DeepFake} Benchmark for Generalizable Forensics},
  year      = {2025},
  note      = {arXiv:2507.18015}
}

@inproceedings{effort,
  author    = {Yan, Zhiyuan and others},
  title     = {Orthogonal Subspace Decomposition for Generalizable {AI}-Generated Image Detection ({Effort})},
  booktitle = {ICML},
  year      = {2025},
  note      = {arXiv:2411.15633}
}

@inproceedings{aide,
  author    = {Yan, Shilin and others},
  title     = {A Sanity Check for {AI}-Generated Image Detection},
  booktitle = {ICLR},
  year      = {2025},
  note      = {arXiv:2406.19435}
}

@inproceedings{genimage,
  author    = {Zhu, Mingjian and others},
  title     = {{GenImage}: A Million-Scale Benchmark for Detecting {AI}-Generated Image},
  booktitle = {NeurIPS Datasets \& Benchmarks},
  year      = {2023},
  note      = {arXiv:2306.08571}
}

@misc{fakeorjpeg,
  author    = {Grommelt, Patrick and others},
  title     = {Fake or {JPEG}? Revealing Common Biases in Generated Image Detection Datasets},
  year      = {2024},
  note      = {arXiv:2403.17608}
}

@misc{aigibench,
  author    = {Li, Ziwen and others},
  title     = {Is Artificial Intelligence Generated Image Detection a Solved Problem? ({AIGIBench})},
  year      = {2025},
  note      = {arXiv:2505.12335}
}

@misc{wildrf,
  author    = {Cavia, Bar and Horwitz, Eliahu and Reiss, Tal and Hoshen, Yedid},
  title     = {Real-Time {Deepfake} Detection in the Real-World},
  year      = {2024},
  note      = {arXiv:2406.09398}
}

@inproceedings{raid,
  author    = {Eddoubi, Hicham and others},
  title     = {{RAID}: A Dataset for Testing the Adversarial Robustness of {AI}-Generated Image Detectors},
  booktitle = {NeurIPS Datasets \& Benchmarks},
  year      = {2025},
  note      = {arXiv:2506.03988}
}

@inproceedings{communityforensics,
  author    = {Park, Jeongsoo and Owens, Andrew},
  title     = {Community Forensics: Using Thousands of Generators to Train Fake Image Detectors},
  booktitle = {CVPR},
  year      = {2025},
  note      = {arXiv:2411.04125}
}

@inproceedings{genvidbench,
  author    = {Ni, Zhenliang and others},
  title     = {{GenVidBench}: A Challenging Benchmark for Detecting {AI}-Generated Video},
  booktitle = {AAAI},
  year      = {2026},
  note      = {arXiv:2501.11340}
}

@misc{fitforpurpose,
  author    = {Lin, Guangyu and Lin, Li and Walker, Christina P. and Schiff, Daniel S. and Hu, Shu},
  title     = {Fit for Purpose? {Deepfake} Detection in the Real World},
  year      = {2025},
  note      = {arXiv:2510.16556}
}

@misc{publictools,
  author    = {Rettinger, Marc and Beaumont, Bertrand and Le-Khac, Nhien-An and Nguyen-Le, Hong-Hanh},
  title     = {How Effective Are Publicly Accessible {Deepfake} Detection Tools? A Comparative Evaluation of Open-Source and Free-to-Use Platforms},
  year      = {2026},
  note      = {arXiv:2603.04456}
}

@misc{decay,
  author    = {Richings, James and Leblanc, Miles and Groves, Iain and Nockles, Victoria},
  title     = {Performance Decay in {Deepfake} Detection: The Limitations of Training on Outdated Data},
  year      = {2025},
  note      = {arXiv:2511.07009}
}

@inproceedings{ffpp,
  author    = {R{\"o}ssler, Andreas and others},
  title     = {{FaceForensics++}: Learning to Detect Manipulated Facial Images},
  booktitle = {ICCV},
  year      = {2019},
  note      = {arXiv:1901.08971}
}

@inproceedings{sbi,
  author    = {Shiohara, Kaede and Yamasaki, Toshihiko},
  title     = {Detecting {Deepfakes} with Self-Blended Images},
  booktitle = {CVPR},
  year      = {2022},
  note      = {arXiv:2204.08376}
}

@inproceedings{caddm,
  author    = {Dong, Shichao and others},
  title     = {Implicit Identity Leakage: The Stumbling Block to Improving {Deepfake} Detection Generalization},
  booktitle = {CVPR},
  year      = {2023},
  note      = {arXiv:2210.14457}
}

@inproceedings{lsda,
  author    = {Yan, Zhiyuan and others},
  title     = {Transcending Forgery Specificity with Latent Space Augmentation for Generalizable {Deepfake} Detection},
  booktitle = {CVPR},
  year      = {2024},
  note      = {arXiv:2311.11278}
}

@inproceedings{unifd,
  author    = {Ojha, Utkarsh and Li, Yuheng and Lee, Yong Jae},
  title     = {Towards Universal Fake Image Detectors that Generalize Across Generative Models},
  booktitle = {CVPR},
  year      = {2023},
  note      = {arXiv:2302.10174}
}

@inproceedings{fatformer,
  author    = {Liu, Huan and others},
  title     = {Forgery-Aware Adaptive Transformer for Generalizable Synthetic Image Detection},
  booktitle = {CVPR},
  year      = {2024},
  note      = {arXiv:2312.16649}
}

@inproceedings{c2pclip,
  author    = {Tan, Chuangchuang and others},
  title     = {{C2P-CLIP}: Injecting Category Common Prompt in {CLIP} to Enhance Generalization in {Deepfake} Detection},
  booktitle = {AAAI},
  year      = {2025},
  note      = {arXiv:2408.09647}
}

@inproceedings{dire,
  author    = {Wang, Zhendong and others},
  title     = {{DIRE} for Diffusion-Generated Image Detection},
  booktitle = {ICCV},
  year      = {2023},
  note      = {arXiv:2303.09295}
}

@inproceedings{drct,
  author    = {Chen, Baoying and others},
  title     = {{DRCT}: Diffusion Reconstruction Contrastive Training towards Universal Detection of Diffusion Generated Images},
  booktitle = {ICML},
  year      = {2024},
  note      = {PMLR 235:7621--7639}
}

@inproceedings{npr,
  author    = {Tan, Chuangchuang and others},
  title     = {Rethinking the Up-Sampling Operations in {CNN}-based Generative Network for Generalizable {Deepfake} Detection},
  booktitle = {CVPR},
  year      = {2024},
  note      = {arXiv:2312.10461}
}

@inproceedings{bfree,
  author    = {Guillaro, Fabrizio and Zingarini, Giada and Usman, Ben and Sud, Avneesh and Cozzolino, Davide and Verdoliva, Luisa},
  title     = {A Bias-Free Training Paradigm for More General {AI}-generated Image Detection},
  booktitle = {CVPR},
  year      = {2025},
  note      = {arXiv:2412.17671}
}

@inproceedings{gpt4mf,
  author    = {Jia, Shan and others},
  title     = {Can {ChatGPT} Detect DeepFakes? A Study of Using Multimodal Large Language Models for Media Forensics},
  booktitle = {CVPR Workshops},
  year      = {2024},
  note      = {arXiv:2403.14077}
}

@inproceedings{fakeshield,
  author    = {Xu, Zhipei and others},
  title     = {{FakeShield}: Explainable Image Forgery Detection and Localization via Multi-modal Large Language Models},
  booktitle = {ICLR},
  year      = {2025},
  note      = {arXiv:2410.02761}
}

@misc{genvideo,
  author    = {Chen, Haoxing and others},
  title     = {{DeMamba}: {AI}-Generated Video Detection on Million-Scale {GenVideo} Benchmark},
  year      = {2024},
  note      = {arXiv:2405.19707}
}

@inproceedings{aigenbench,
  author    = {Pellegrini, Lorenzo and others},
  title     = {{AI-GenBench}: A New Ongoing Benchmark for {AI}-Generated Image Detection},
  booktitle = {Verimedia Workshop @ IJCNN},
  year      = {2025},
  note      = {arXiv:2504.20865}
}

@techreport{sotf,
  author      = {{BitMind Research Team}},
  title       = {Survival of the Fittest Detectors: A Decentralized Framework for Evolving {Deepfake} Detection},
  institution = {BitMind Research},
  year        = {2024},
  note        = {\url{https://assets.bitmindlabs.ai/pdfs/survival-of-the-fittest-detectors.pdf}}
}

@inproceedings{cnnspot,
  author    = {Wang, Sheng-Yu and Wang, Oliver and Zhang, Richard and Owens, Andrew and Efros, Alexei A.},
  title     = {{CNN}-Generated Images Are Surprisingly Easy to Spot\ldots for Now},
  booktitle = {CVPR},
  year      = {2020},
  note      = {arXiv:1912.11035}
}

@inproceedings{red140,
  author    = {Guo, Xiao and Asnani, Vishal and Liu, Sijia and Liu, Xiaoming},
  title     = {Tracing Hyperparameter Dependencies for Model Parsing via Learnable Graph Pooling Network},
  booktitle = {NeurIPS},
  year      = {2024},
  note      = {arXiv:2312.02224}
}

@inproceedings{hdit,
  author    = {Crowson, Katherine and Baumann, Stefan Andreas and Birch, Alex and Abraham, Tanishq Mathew and Kaplan, Daniel Z. and Shippole, Enrico},
  title     = {Scalable High-Resolution Pixel-Space Image Synthesis with {Hourglass} Diffusion Transformers},
  booktitle = {ICML},
  year      = {2024},
  note      = {arXiv:2401.11605}
}

@inproceedings{clip,
  author    = {Radford, Alec and Kim, Jong Wook and Hallacy, Chris and others},
  title     = {Learning Transferable Visual Models From Natural Language Supervision},
  booktitle = {ICML},
  year      = {2021},
  note      = {arXiv:2103.00020}
}

@inproceedings{convnext,
  author    = {Liu, Zhuang and Mao, Hanzi and Wu, Chao-Yuan and Feichtenhofer, Christoph and Darrell, Trevor and Xie, Saining},
  title     = {A ConvNet for the 2020s},
  booktitle = {CVPR},
  year      = {2022},
  note      = {arXiv:2201.03545}
}

@article{eva02,
  author    = {Fang, Yuxin and Sun, Quan and Wang, Xinggang and Huang, Tiejun and Wang, Xinlong and Cao, Yue},
  title     = {{EVA-02}: A Visual Representation for Neon Genesis},
  journal   = {Image and Vision Computing},
  volume    = {149},
  year      = {2024},
  note      = {arXiv:2303.11331}
}

@misc{dinov3,
  author    = {Sim{\'e}oni, Oriane and Vo, Huy V. and Seitzer, Maximilian and Baldassarre, Federico and others},
  title     = {{DINOv3}},
  year      = {2025},
  note      = {arXiv:2508.10104}
}

@misc{vjepa2,
  author    = {Assran, Mido and others},
  title     = {{V-JEPA}~2: Self-Supervised Video Models Enable Understanding, Prediction and Planning},
  year      = {2025},
  note      = {arXiv:2506.09985}
}

@article{mtcnn,
  author    = {Zhang, Kaipeng and Zhang, Zhanpeng and Li, Zhifeng and Qiao, Yu},
  title     = {Joint Face Detection and Alignment Using Multi-task Cascaded Convolutional Networks},
  journal   = {IEEE Signal Processing Letters},
  volume    = {23},
  number    = {10},
  pages     = {1499--1503},
  year      = {2016},
  note      = {arXiv:1604.02878}
}

@inproceedings{gpen,
  author    = {Yang, Tao and Ren, Peiran and Xie, Xuansong and Zhang, Lei},
  title     = {{GAN} Prior Embedded Network for Blind Face Restoration in the Wild},
  booktitle = {CVPR},
  year      = {2021},
  note      = {arXiv:2105.06070}
}

@inproceedings{simswap,
  author    = {Chen, Renwang and Chen, Xuanhong and Ni, Bingbing and Ge, Yanhao},
  title     = {{SimSwap}: An Efficient Framework For High Fidelity Face Swapping},
  booktitle = {ACM Multimedia},
  year      = {2020},
  note      = {arXiv:2106.06340}
}

@article{delong,
  author    = {DeLong, Elizabeth R. and DeLong, David M. and Clarke-Pearson, Daniel L.},
  title     = {Comparing the Areas under Two or More Correlated Receiver Operating Characteristic Curves: A Nonparametric Approach},
  journal   = {Biometrics},
  volume    = {44},
  number    = {3},
  pages     = {837--845},
  year      = {1988}
}
}
\end{document}